\documentclass[conference]{IEEEtran}

\usepackage[utf8]{inputenc}
\usepackage[numbers]{natbib}
\usepackage{graphicx}
\usepackage{mathtools}
\usepackage{amsmath}
\usepackage{booktabs}
\usepackage{xcolor}
\usepackage{url}
\usepackage[ruled,vlined]{algorithm2e}
\usepackage{subfig}
\usepackage{multirow}
\usepackage{amssymb}

\begin{document}

\title{CASPER: Cognitive Architecture for Social Perception and Engagement in Robots}
\author{\IEEEauthorblockN{Samuele Vinanzi* and
Angelo Cangelosi}

\IEEEauthorblockA{Cognitive Robotics Lab, The University of Manchester, Manchester, United Kingdom\\ \\ *Correspondence: \(samuele.vinanzi@manchester.ac.uk\)}
}

\maketitle

\begin{abstract}
Our world is being increasingly pervaded by intelligent robots with varying degrees of autonomy. To seamlessly integrate themselves in our society, these machines should possess the ability to navigate the complexities of our daily routines even in the absence of a human's direct input. In other words, we want these robots to understand the intentions of their partners with the purpose of predicting the best way to help them. In this paper, we present CASPER (Cognitive Architecture for Social Perception and Engagement in Robots): a symbolic cognitive architecture that uses qualitative spatial reasoning to anticipate the pursued goal of another agent and to calculate the best collaborative behavior. This is performed through an ensemble of parallel processes that model a low-level action recognition and a high-level goal understanding, both of which are formally verified. We have tested this architecture in a simulated kitchen environment and the results we have collected show that the robot is able to both recognize an ongoing goal and to properly collaborate towards its achievement. This demonstrates a new use of Qualitative Spatial Relations applied to the problem of intention reading in the domain of human-robot interaction.


\end{abstract}


\section{Introduction}
\label{sec:introduction}

Autonomous robots are increasingly present in our everyday life. Once limited to research laboratories and industrial settings, they now frequently inhabit our living spaces and interact with us during our day. This new generation of intelligent machines, categorized under the umbrella term of ``social robotics'', is expected to navigate a complex and uncertain landscape made of human beliefs, desires, intentions and social norms. Additionally, it is desirable for these agents to be able to act autonomously, that means without direct input from their human partners.

\begin{figure}[t]
	\begin{flushright}
		\includegraphics[width=\linewidth]{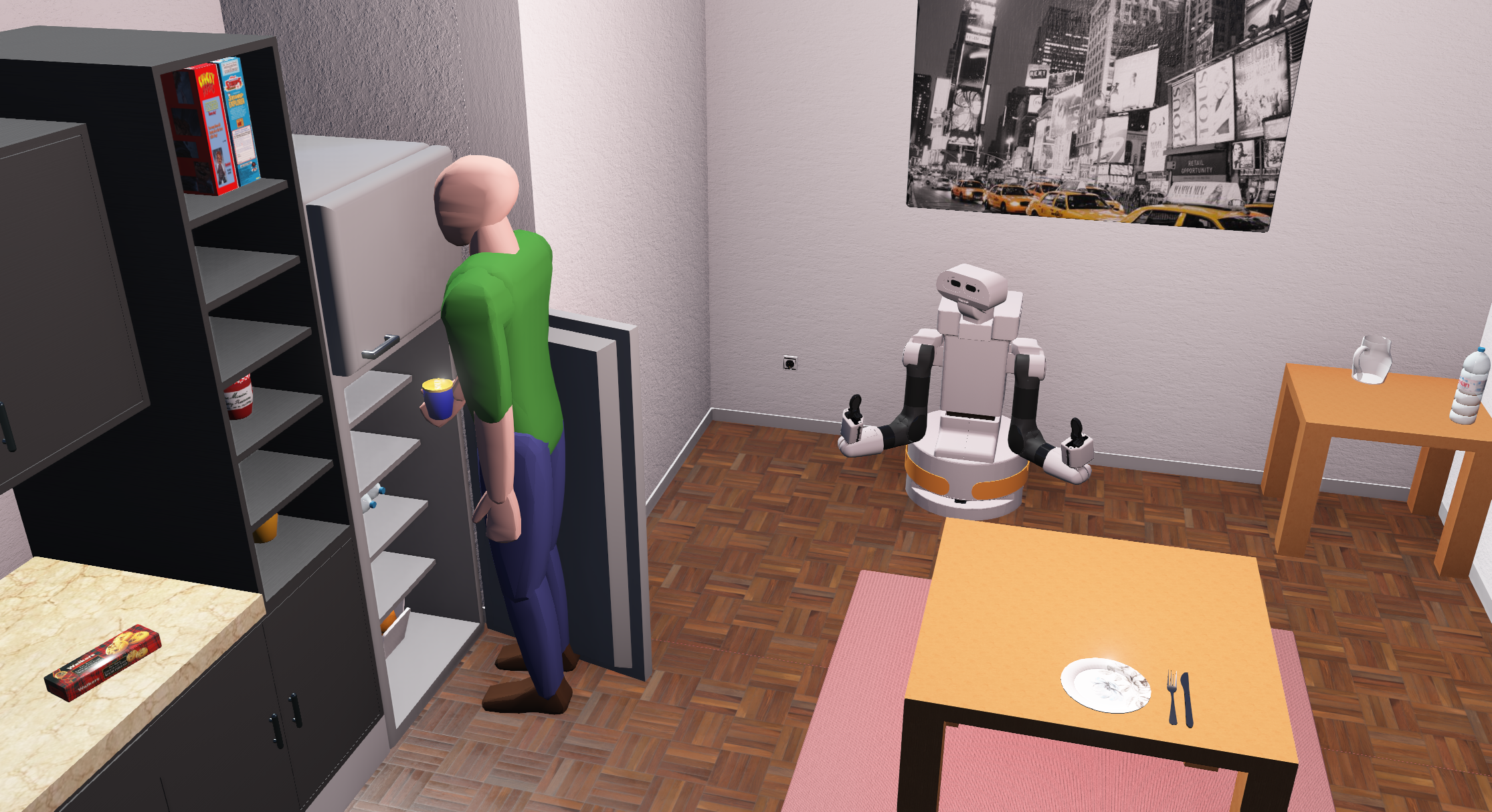}
	\end{flushright}
	\caption{The simulated robot equipped with CASPER observes the actions of another agent (in this case, a simulated human) in order to predict their goal and the best collaborative plan.}
	\label{fig:intro}
\end{figure}

In order to integrate these robots into our society, it is paramount for them to be endowed with the same set of cognitive and mental skills that regulate the way in which we, as people, interact with other agents. One of the most fundamental of such skills is known as ``intention reading'' and represents the capacity to understand the implicit goal that is driving the actions of another agent \cite{vinanzi_intentions}. By making appropriate use of this ability, we can allow a social robot to use its observations of another agent to extrapolate their underlying goal, reconstruct their expected future actions and, finally, determine how and when to enact collaborative behavior.

This paper presents CASPER (Cognitive Architecture for Social Perception and Engagement in Robots), a platform-independent cognitive architecture that uses a mixture of symbolic and data-driven artificial intelligence methodologies to perform intention reading and collaboration in a human-robot interaction (HRI) scenario. ``Social perception'' refers to the act of identifying and using social cues to make judgments about others, while with the term ``engagement'' we highlight the system's ability of translating this knowledge into practical involvement and interaction.

This system lays its foundations on the use of Qualitative Spatial Relations (QSRs) that describe how the observed partner moves inside the environment with respect to the Objects of Interest (OOIs). CASPER analyzes the temporal evolution of these descriptors to predict future actions, which are subsequently validated by a knowledge-base system and processed by a probabilistic plan recognizer. This data-driven approach to recognition is complemented by a top-down decision-making strategy to generate assistive activity.

Our contributions to the field include the following:

\begin{itemize}
	\item A new cognitive architecture that implements intention reading and collaborative behavior capabilities for human-robot interaction, the need of which arises from the scarcity of such models in the current literature \cite{kotseruba202040}.
	\item To the best of our knowledge, CASPER represents one of the first attempts to introduce the use of QSR descriptors to perform efficient and easily generalizable intention reading for embodied robots. This paper presents a proof of concept on the untapped potential of these mathematical tools and tries to promote their use in future cognitive architectures for action and goal recognition.
	\item On a technical level, we present a collection of novel algorithms for social perception and reasoning that take inspiration from well-established psychology and cognitive science principles.
	\item We provide the first case study for this cognitive architecture: a demonstration of its possible implementation to solve a collaborative task based in a kitchen environment.
\end{itemize}

Our long-term goal is to implement CASPER into an heterogeneous multi-agent teaming scenario, where a team of distributed agents, both humans and robots of different kind, are engaged in joint action to achieve a common goal. In order to do so, this paper builds the foundations of this cognitive architecture and presents a first case study involving a dyadic interaction that takes place inside a kitchen. Within this specific application of CASPER, the robot is expected to observe its partner, identify the goal that is driving their actions and calculate the best way to assist with their task. Our experiments carried out in simulation (Figure \ref{fig:intro}) show that the robot using this cognitive architecture is able not only to accurately predict the partner's goals before they are achieved, but also to formulate appropriate collaborative decision-making. This allowed the human to silently and implicitly delegate part of the task to their artificial companion.

The organization of the paper is as follows: Section \ref{sec:previous} offers a general background on artificial cognitive architectures and intention reading, other than introducing the notion of QSRs. Section \ref{sec:method} discusses CASPER's design specifications and algorithmic details. Section \ref{sec:experiments} covers the implementation of the general CASPER architecture to the specific example of the simulated kitchen environment. Section \ref{sec:results} discusses the performance of the system on the selected case study. Finally, Section \ref{sec:conclusion} concludes and highlights possible future directions.

\section{Previous Work}
\label{sec:previous}

\subsection{Cognitive Architectures for Robots}
\label{sub:coro}
Cognitive Robotics is a discipline that lies at the intersection of robotics and cognitive science, which is the scientific study of the mind and its processes such as perception, attention, anticipation, planning, memory, learning, and reasoning. It has been defined as ``the field that combines insights and methods from artificial intelligence, as well as cognitive and biological sciences, to robotics'' \cite{cangelosi2022cognitive}. This definition highlights the interdisciplinary nature of this approach, which takes inputs from linguistics, psychology, neuroscience, philosophy, computer science and anthropology. Its aim is to create intelligent robots which are endowed with the same set of mental skills as a human being.

Cognitive science views the mind as an information processor and studies the operations through which perceptual stimulus are combined to obtain higher-level mental functions \cite{friedenberg2021cognitive}. These principles are easily transferable to an embodied robotic platform that can implement the same functions despite the difference in the underlying structure (a brain versus a computer). This is done by designing what is known as an ``artificial cognitive architecture'': a computational system which instantiates one or more cognitive theories using artificial intelligence methodologies in an attempt to model the human mind.

A recent review has estimated the existence of around 300 cognitive architectures in the current literature \cite{kotseruba202040}. The vast majority of them specialize on modeling particular aspects of cognition such as attention \cite{tsotsos2017attention}, emotion \cite{faghihi2011emotional} or problem solving \cite{epstein2004metaknowledge}, while only a fraction aims to achieve Artificial General Intelligence. The latter case includes some of the most famous architectures in the current literature, such as ACT-R \cite{anderson2013architecture}, Soar \cite{laird1987soar}, LIDA \cite{snaider2011lida} and NARS \cite{wang1995non}. These are all implemented as general frameworks that can be deployed to specific use cases, including applications in robotics \cite{chella2019cognitive}. For example, ACT-R is written as a Common Lisp interpreter and its applications come in the form of scripts in the ACT-R language. Despite their purpose, every cognitive architectures usually models one or more aspects of cognition such as perception, attention, action selection, memory, learning and reasoning \cite{kotseruba202040}. 

CASPER belongs to the category of more specialized architectures and focuses on modeling specifically human-robot collaboration (HRC) mental capabilities and, in particular, intention reading. Other cognitive systems that belong to the same class make use of different techniques such as unsupervised clustering of human postures \cite{vinanzi_intentions} and artificial mirror neuron networks \cite{han2010human}.

\subsection{Human-Robot Collaboration}

The human species' success is ascribable to its ability to collaborate with others to obtain otherwise inaccessible goals. This instinct towards cooperation is shared by our close relatives in the animal kingdom, which however use it as a means to achieve a purpose, rather than being intrinsically motivated in pursuing it \cite{melis2013evolutionary}. Given the importance that collaboration has for humans, it seems natural to try and transpose this skill to the autonomous intelligent machines that we are designing as everyday companions.

HRC is a branch of HRI which studies the best ways to ensure a safe and effective interaction between humans and robots engaged in joint tasks with common goals: we call this a ``team''. The vast majority of works in this field's literature focus on industrial settings, where robotic arms such as Sawyer or Kuka robots offer assistance in some kind of assembly task \cite{el2017design, wang2019symbiotic, tsarouchi2017human}. Many of these studies deal with the scheduling and subdivision of tasks between the two agents. Hoffman et al. \citep{hoffman2004collaboration} argue that collaborative robots (or ``cobots'') should possess communication mechanisms in order to both understand humans and to inform them about
their own goals, and in so doing maintain a set of shared beliefs which support the execution
of a joint plan. In fact, many HRC models focus on direct verbal cooperation and implement dialogue managers \cite{bluethmann2003robonaut, lallee2010human, pineau2003towards}. More recently, a number of studies have adopted Machine Learning methodologies (sometimes encased in artificial cognitive architectures) to learn and modulate the robot's response to human tasks \cite{semeraro}.

\subsection{Intention Reading}

On a psychological point of view, one of the main cognitive skills that enable social and collaborative attitudes in humans is known as ``intention reading'' \cite{woodward2009emergence}: this is the ability to understand the driving goal of another agent by the observation of their physical clues. This is possible because we don't understand the behavior
of others as a series of unrelated motions through space, but rather as sets of goal-directed
actions \cite{malle2001intentions}. Intention reading is a fundamental skill to implement in a collaborative intelligence because an agent has to first understand what their partner's goal is before knowing how to offer its assistance.

Balwdin et al. \citep{baldwin2001discerning} state that humans process continuous actions as streams of hierarchical relations that link low-level intentions (such as grasping a plate, bringing it to the sink and opening the tap) to high-level intentions (to wash the dishes or clean the kitchen). Additionally, they state that adults more reliably identify the higer-level goals based on some actions that are understood to be more crucial than others (for example, they point to the fact that scrubbing a plate is a stronger signal for the intention to wash the dishes than the equally necessary but less central action of turning on the water). Some of the most important social cues that are used for this purpose are biological motion and gaze direction \cite{tomasello_carpenter_call_behne_moll_2005}.

On a computational perspective, there have been many attempts to develop intention reading models for HRI. A notorious experiment by Dominey and Warneken \cite{dominey2011basis} investigates the shared intentionality in a turn-taking game between a human and a robotic arm, where the artificial agent would build a representation of the shared plan and subdivide the actions between itself and its partner. Duarte et al. \cite{duarte2018action} perform action anticipation exploiting several social cues such as saccadic eye movement, gaze directing and arm movements processed through a Gaussian Mixture Model. Bien et al. \cite{bien2005intention} developed a system that analyzes posture and movements in elderly people and tries to decode the inner intentions they are driven by. Other relevant researches involve the use of dynamic
Bayesian networks \cite{tahboub2006intelligent}, self-organizing maps \cite{buonamente2013recognizing}, first order logic \cite{jansen2006computational} or the imitation of the biological mirror neuron system \cite{oztop2006mirror}. An interesting approach has been adopted by Granada et al. \cite{granada1995hybrid}, who divide the intention reading task on a low-level action recognition paired with a high-level goal understanding and execute the task combining a convolutional neural network with a symbolic plan recognizer.

Previous works by the authors \cite{vinanzi2021collaborative, vinanzi_goals, vinanzi_intentions} have explored the use of social cues and psychological theories to develop intention reading capabilities in humanoid robots. In these works, we analyzed kinematic signals (body posture, head gaze) to infer the human's current goal using clustering algorithms paired with probabilistic modeling. Our interest is shifting from dyadic interactions to heterogeneous multi-agent environments in which the Team Goal is shared among sub-teams distributed along a structured world. For this reason, we felt the need to adopt a higher-level approach and shift our focus from subtle kinematic signals to qualitative relations between the entities in the environment. This work lays the foundations of CASPER, which will be in the future applied to a complex scenario like the one we have just described.

\subsection{Qualitative Spatial Relations}
Most mathematical and engineering modeling of phenomena rely on \emph{quantitative} representations: measurements made on standard units that define physical properties. This is not how human commonsense generally works: in order to process information in a timely and efficient manner, we tend to discard unnecessary details, reducing the concept of space into coarse categories. For example, in our everyday life we would say that something is ``far away'' without specifying the exact number of meters. This is an example of \emph{qualitative} spatial reasoning.

QSRs are tools that allow commonsense reasoning about space and time using qualitative relations for different spatial aspects such as topology, direction and position \cite{Moratz2017}. In other words, QSRs enable an agent to reason about actions such as ``the human is approaching the fridge'' without having to maintain precise metric information on the positions of both the actor and the reference point (in this example, the human and the fridge). Such a representation does not depend on factors such as the starting positions of the actor and reference, the speed of movement or its exact trajectory and so it is easily generalizable.

QSRs have been used for a broad range of artificial intelligence applications, such as human activity learning \cite{duckworth2019unsupervised} and monitoring \cite{behera2012workflow}, language learning \cite{dubba2014grounding}, imitation learning \cite{young2015learning} and even to encode spatial structure features for artificial neural networks \cite{krishnaswamy2019combining}.

The use of QSRs for intention reading purposes is, to the best of our knowledge, novel. Many papers that make use of them focus on action recognition, which is an important but non exhaustive step in the process of understanding the underlying goal that is driving an agent's behavior. One of the main scientific contributions of this paper is to demonstrate that QSRs can be effectively used in this domain, since they are more prone to generalization than the features usually employed by intention reading models (such as motion trajectories or posture keypoints).

\section{Proposed Method}
\label{sec:method}

\begin{figure*}[]
	\includegraphics[width=\textwidth]{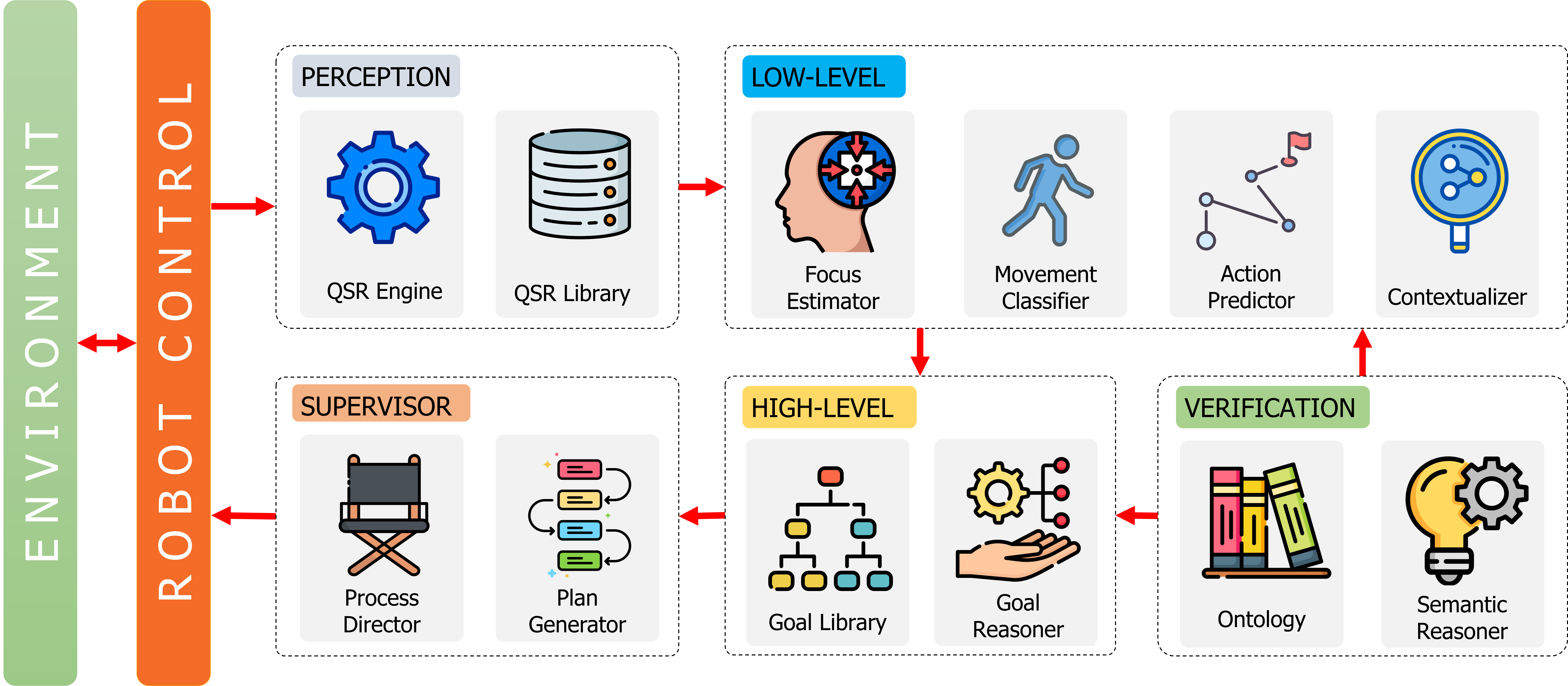}
	\caption{Overview of the proposed system. CASPER is composed of several parallel processes that interact with each other in a joint effort to decode an agent's intention and to formulate an appropriate response. The Perception module transforms visual observations into QSRs. The latter are used by the Low-Level process to predict the actions that are being performed in the environment and passes this information to the High-Level component, which tries to match them against the plan library to infer the pursued goal. A knowledge base, enveloped in the Verification module, ensures the step-by-step soundness of these predictions. Finally, the Supervisor coordinates all the other processes, collects the results and composes a collaborative plan that will be executed by the robot.}
	\label{fig:architecture}
\end{figure*}

The purpose of our work is the development of an artificial cognitive architecture that will allow an autonomous social robot to observe the actions of a partner (be it a human or a humanoid), understand their underlying goal and collaborate on the ongoing task. Our design choices are inspired by the low- and high-level subdivision of actions and goals theorized by psychologists and use biologically plausible inputs \cite{tomasello_carpenter_call_behne_moll_2005}. We allow this model to generate an appropriate assistive response by leveraging on the prediction of the current goal, the observed actions and the incomplete part of the plan. The overall system is depicted in Figure \ref{fig:architecture}. The architecture is composed of several parallel processes which gather, elaborate and share data between them.

\subsection{Goal and Intention Representation}
\label{sub:goal}

\begin{figure}[t]
	\includegraphics[width=\linewidth]{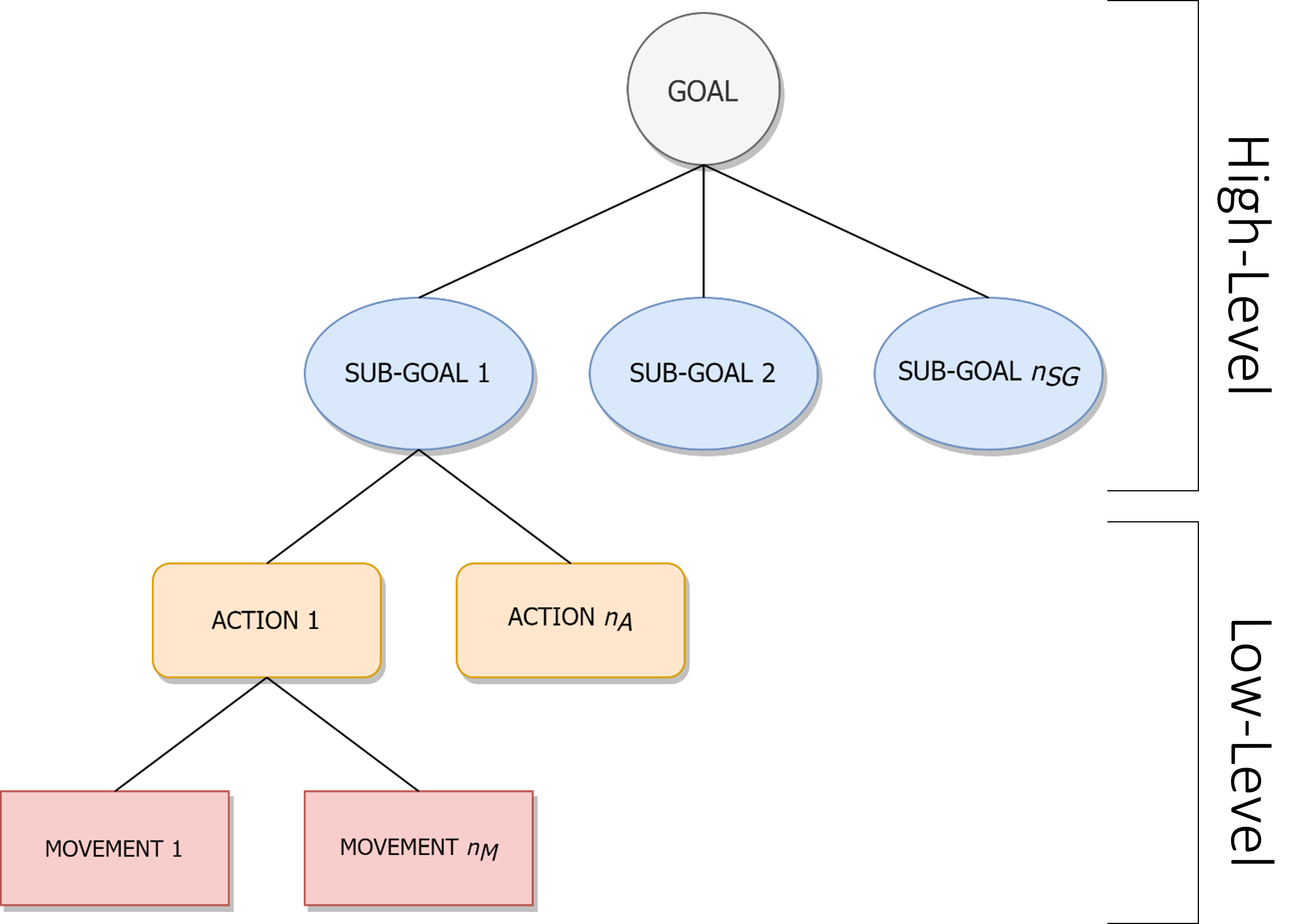}
	\caption{A plan in CASPER. Each goal is formed by a temporally ordered set of sub-elements with varying levels of abstraction. This structure is used for both intention reading and collaborative behavior generation.}
	\label{fig:goal}
\end{figure}

Before giving a detailed description of each component showed in Figure \ref{fig:architecture}, it is worth explaining how this cognitive architecture represents intentions. Within CASPER, every goal is described by a Plan: a sequence of events that have to be executed in order for the task to be accomplished. Figure \ref{fig:goal} describes this structure as a non-binary tree. The main \emph{goal} is formed by a collection of \emph{sub-goals}, each of which is achievable by a sequence of \emph{actions}. The latter, in turn, are composed by a set of \emph{movements}. The nodes of the Plan are temporally ordered from left to right.

The nodes represented as ellipses in Figure \ref{fig:goal} are the abstract, conceptual representations of the plan and fall under the domain of the High-Level module (Section \ref{sub:highlevel}), while the ones drawn as rectangles are the ones that can be directly observed by the Low-Level module (Section \ref{sub:lowlevel}). The hierarchical organization of goals is inspired by cognitive science \cite{friedenberg2021cognitive}.

This data structure is used both when reading the partner's intention (bottom-up, data-driven inference) and when generating an appropriate collaborative behavior for the robot (top-down, conceptually-driven inference).

\subsection{Environment and Robot Control}
\label{sub:enviro}

This architecture assumes the existence of a robot $r$ which is observing another agent $a$ performing actions within an environment $E$. The latter will contain a set $O = \{o_1, o_2, ... , o_n \}$ of OOIs which can be interacted with. For example, possible OOIs in a kitchen would be pieces of cutlery and food items, but not structural elements such as walls and floors.\\

We have designed this cognitive architecture to be platform-independent. This means that any robot with basic vision capabilities can be programmed to interact with CASPER. Of course, the physical limitations of the chosen robot define how it will be able to interact with the world and assist its partner, but this information can be easily encapsulated in the knowledge base (see Section \ref{sub:verification}).

\subsection{Perception}
\label{sub:perception}

The Perception module interacts directly with the Robot Control system and is used to produce a qualitative description of the relations between the observed agent and the OOIs in the environment. The agent will periodically update this component by transmitting a World State: a dictionary that, at every timestep, records the absolute coordinates of each OOIs calculated from the robot's visual sensors (such as RGB or RGBD cameras). This World State is then processed by the QSR Engine to obtain the qualitative spatial descriptors of the scene at the current timestep. For our purposes, we chose to use QSRlib \cite{gatsoulis2016qsrlib}, an open-source software library designed to calculate QSRs from a scene description.

The QSRs we calculate from our World State are the following:

\begin{itemize}
	\item Qualitative Distance Calculus (QDC) \cite{clementini1997qualitative}.
	\item Qualitative Trajectory Calculus (QTC) \cite{delafontaine2011implementing}.
	\item Moving or Stationary (MOS).
	\item Holding Object (HOLD).\\
\end{itemize}

\textbf{QDC}: defines the qualitative Euclidean distance between two entities in the scene, which in our case are $a$ and $o_i \in O$. On an intuitive level, this QSR describes how close the agent is to the OOIs under consideration. The thresholds are parameterized within QSRlib and for our purposes we have defined the following: `touch' [0-0.6m], `near' (0.6-2m], `medium' (2-3m], `far' (3-5m] and `ignore' for distances greater that 5m.

\textbf{QTC}: represents the relative motion between a set of moving point objects having a free trajectory in an $n$-dimensional space. The current literature contains several variations of this descriptor \cite{van2004representing}, out of which we chose QTC$_{B11}$. The latter involves two points (which for our purposes will be $a$ and $o_i$) and makes use of the Euclidean distance calculated on the reference line that connects them. Because of the nature of this QSR, it can only be calculated over two distinct timesteps. The results of this calculation can be either: $a$ is stationary with respect to $o_i$ (represented by the symbol $0$), $a$ is moving towards $o_i$ ($-$) or $a$ is moving away from $o_i$ ($+$).

\textbf{MOS}: a unary QSR that describes whether the entity is in motion or stationary between two different timesteps.

\textbf{HOLD}: this unary descriptor indicates whether the agent $a$ is holding an object in one of their hands or not.\\

After having produced a QSR description of the current timestep, the Perception module stores it in a local library, which acts as the system's sensory memory. The latter contains all the time-ordered QSRs calculated since the beginning of the activity and it can be accessed on request by other processes.

\subsection{Low-Level Action Recognition}
\label{sub:lowlevel}

This component of the cognitive architecture is in charge of predicting the first elements of the data-driven inference as shown in Figure \ref{fig:goal}: movements and actions. The main idea behind this module is to incrementally aggregate and refine data: a set of QSRs will be classified as a movement and a set of movements will define an action.

\subsubsection{Focus Estimation}
\label{subsub:focus}

\begin{table}
	\centering
	\caption{Encodings for the QSRs used in the Focus Estimator.}
	\begin{tabular}{|c|c|}
		\hline
		\textbf{QDC} & \textbf{Encoding} \\ \hline
		touch      & 0.5               \\ \hline
		near       & 0.25              \\ \hline
		medium     & 0.125             \\ \hline
		far        & 0                 \\ \hline
	\end{tabular}
	\quad
	\begin{tabular}{|c|c|}
		\hline
		\textbf{QTC} & \textbf{Encoding} \\ \hline
		0          & 0.5               \\ \hline
		-          & 0.25              \\ \hline
		+          & 0                 \\ \hline
	\end{tabular}
\label{tab:encodings}
\end{table}

If there are $n$ OOIs in the environment, the Low-Level would be able to perform $n$ distinct classifications, one for each OOI. This means that, at every timestep, CASPER would be able to identify $n$ different movements. In reality, we know that the partner is an intentional agent which is performing one movement at a time, directed towards a specific entity in the scene which we call the ``target''.

The challenge, then, is to identify the target using perceptual information from the observed scene. Given the inherent uncertainty of this task, it seems natural to solve it through the use of probabilistic models. Our algorithm assigns a score $S$ to each OOI based on the following equation, which envelops the use of motion and gaze as perceptual inputs to the intention reading cognitive functions theorized by Tomasello et al. \cite{tomasello_carpenter_call_behne_moll_2005}:

\begin{equation}
	S(o_i) = \frac{w_{QDC} \cdot QDC(o_i) + w_{QTC} \cdot QTC(o_i)}{1 + \theta}
	\label{eq:focus}
\end{equation}

for $o_i \in O$. In the above equation, $QDC(o_i)$ and $QTC(o_i)$ represent the QDC and QTC QSRs calculated on OOI $o_i$. These are categorical variables, so they need to be encoded into numerical values using the conversion shown in Table \ref{tab:encodings}. $w_{QDC}$ and $w_{QTC}$ are the positive weights that we assign to these components, with $w_{QDC} + w_{QTC} = 1$.

Given an uniform weight distribution, the numerator in Equation \ref{eq:focus} is maximized when the agent is maintaining touching distance with the object. On the contrary, it is minimized when the agent is at a far distance, walking away from it. This value is scaled by $\theta$, which represents the angle between the agent's heading and the reference line connecting them with the OOI. Intuitively, the denominator penalizes OOIs which are not in the field of view of the agent.\\

After calculating the attention score for every OOI, these are normalized into probabilities. To win the competition and be elected as the target, an element must both possess the maximum score and surpass the threshold $\tau = 0.5$. To eliminate any possible noise in the prediction that would affect the processing chain, this item is not forwarded as it is, instead it is inserted into a sliding window of size $w = 4$ which allows the system to select the target as a measure of central tendency. Any OOI that at any time occupies the majority of the sliding window slots is declared to be the current target of the observed agent's focus.

The Focus Estimator keeps also track of the second-highest scoring OOI, processing it independently using the same procedure described above. This element, when it exists, is understood to be the ``destination'', which will be later used to contextualize the agent's action. For example, if the agent is transporting an empty glass towards a bottle, then they are probably going to fill it and have a drink, conversely if the destination is the sink, then they will likely going to wash it.

In case a tie occurs between the OOIs, if one of them was previously declared as a target it will maintain its status, while the other one will be regarded as the destination.\\

\subsubsection{Movement Classification}
\label{subsub:movement}

\begin{figure}[t]
	\includegraphics[width=\linewidth]{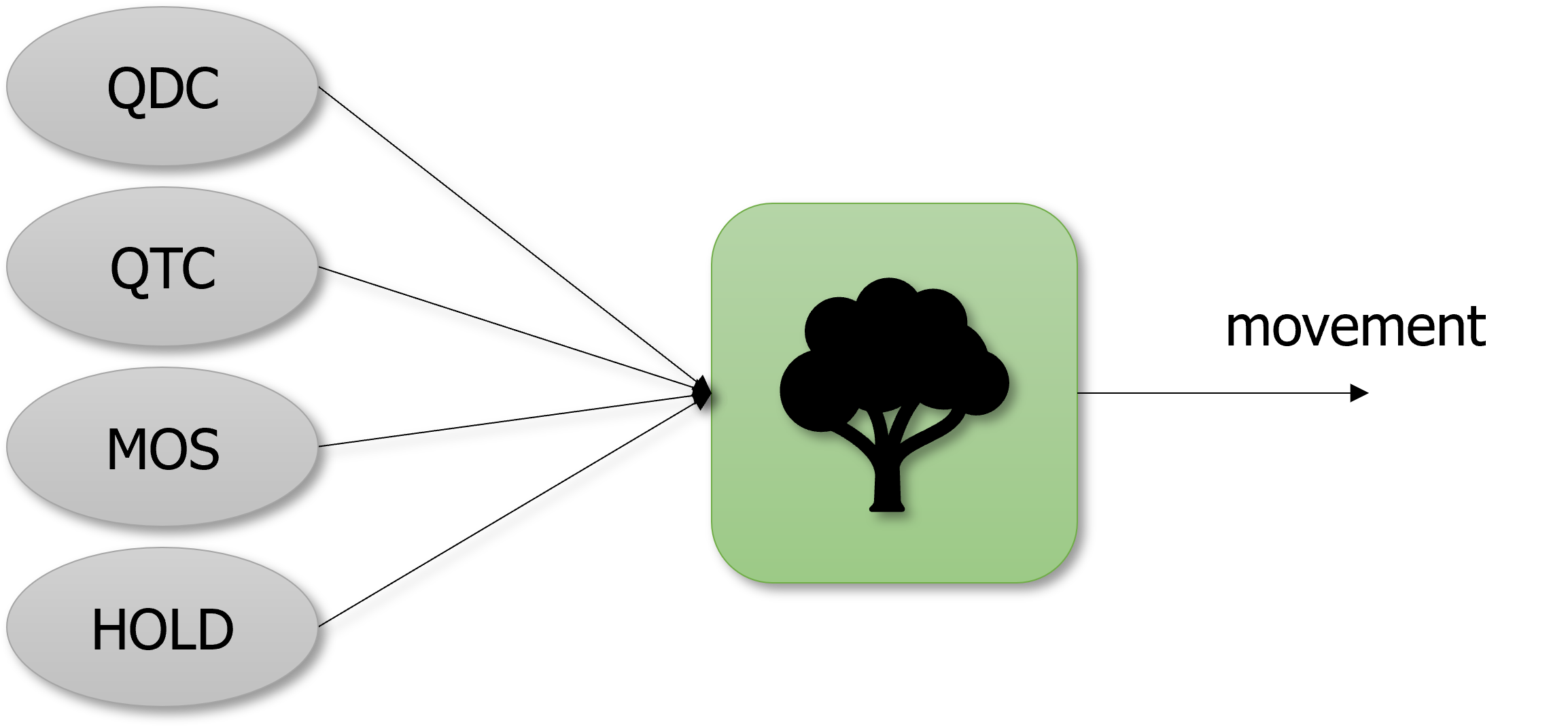}
	\caption{The Decision Tree maps a set of QSRs into a movement.}
	\label{fig:tree}
\end{figure}

Once the Focus Estimator has identified the partner's target, it is possible to proceed with the movement prediction with respect to the inferred OOI. For this purpose, we make use of a symbolic data-driven model: specifically, a Decision Tree. We have chosen this model because it fits well for our purpose, that is to form a mapping from a set of QSRs to a domain of discrete movements, each representing a motion that the observed agent is performing in a single timestep. 

A graphical representation of this process can be viewed in Figure \ref{fig:tree}.\\

\subsubsection{Action Prediction}
\label{subsub:action}

Through the procedures described in the previous sections, CASPER is able to determine the movement performed by the agent at each timestep, providing the robot with an instantaneous information of what is happening in a single unit of time. To fully understand the behavior of the partner, however, we need to analyze the temporal evolution of these movements: we call this an ``action''.

To represent the composition of each action, we use a Markov-chain Finite State Machine (FSM). This model describes a process in which the transition to a state at time $t+1$ is probabilistic rather than deterministic and depends only on the state at time $t$. Then, we combine these FSMs in an ensemble which is used to classify a sequence of observations. Every time the Movement Classifier generates new data, this is initially filtered such that only transitions between different states are considered. This technique, already implemented in other activity recognition and prediction models \cite{vinanzi_intentions, manzi2017human}, allows the system to be time-invariant: this means that the speed at which the action is performed does not influence its representation.

Every time a new filtered observation is detected, it is queued up with the previous ones. The ensemble samples each of its FSMs using the initial observation as the first state of the chain. Thereafter, it calculates the similarity between the ordered sequence of observations and the generated samples using the Ratcliff-Obershelp Pattern Recognition algorithm \cite{ratcliff1988pattern}. Each FSM is assigned a score in $[0,1]$ based on this metric. An action is predicted when there is a clear winner between the models, but only if the score surpasses a certain threshold: this allows the system to be more robust to transient effects that will create noise in the inference.\\

Actions might be ambiguous and require contextualization based on the location within the environment in which they are performed. When this is the case, we have opted to use a simple lookup table in which we use information from the destination (see Section \ref{subsub:focus}) to disambiguate actions.

\subsection{High-Level Goal Prediction}
\label{sub:highlevel}

\begin{figure}[t]
	\centering
	\includegraphics[width=\linewidth]{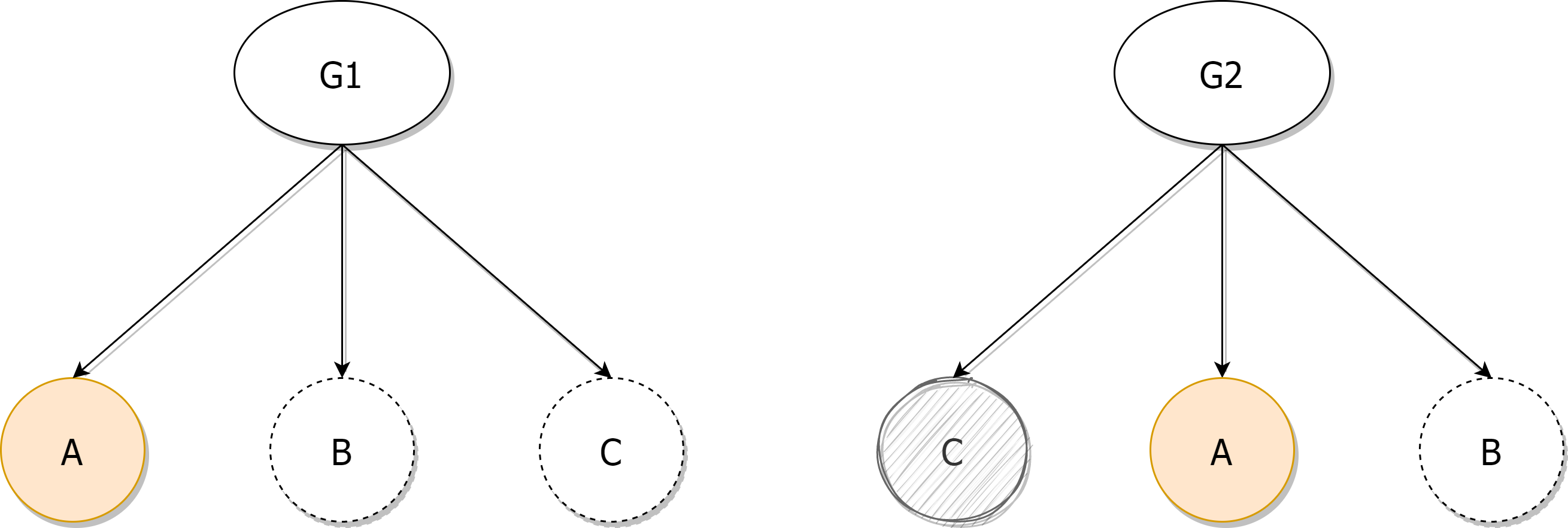}
	\caption{A visual demonstration of the Plan Library's scoring system. These trees represent two plans for two distinct goals with a single observation $\hat{\sigma_1} = A$. The non-root nodes are drawn differently based on their status: filled if observed, dashed if unobserved and textured if missed. In this example, $P({G1} | \hat{\sigma}) = 0.59$ and $P({G2} | \hat{\sigma}) = 0.39$, so $G1$ is considered the best explanation.}
	\label{fig:score}
\end{figure}

The purpose of this component is to form a computational representation of the plan structure described in Figure \ref{fig:goal} on which it is possible to execute inference and reasoning. We model this data structure as a non-binary tree where the root represents the goal and each terminal node is a possible action (derived from sequences of movements as described in Section \ref{sub:lowlevel}).

The Plan Library $L$ is then defined as:

\begin{equation}
	L = \{ \Sigma, NT, G, T \}
	\label{eq:pl}
\end{equation}

Where $\Sigma$ is a set of terminal symbols that represent the observable actions, $NT$ are the non-terminal symbols which stand as sub-goals, $G$ is the set of goals and $T$ are the trees that describe the ordered production rules that compose the plans.

During the intention reading process, the Low-Level module will produce a serialized set of observed actions $\hat{\sigma} = \{\hat{\sigma}_1, ... , \hat{\sigma}_n\} \in \Sigma$ that have to be matched against the available plans $T$ in the Plan Library in order to infer which goal $g \in G$ is driving the observable actions of the agent. We work on the assumption that $\hat{\sigma}$ is temporally ordered, which means that $\hat{\sigma}_1$ happens and is observed before $\hat{\sigma}_2$ occurs.

The intuition behind our design is the following: using Occam's razor principle, the plan that better describes the data is the one that more simply fits the observations and that leaves less gaps in the explanation (intended as nodes that should have been observed but are not present in $\hat{\sigma}$). 

The probability that goal $g$ is generating the observations $\hat{\sigma}$ is:

\begin{equation}
	P(g | \hat{\sigma}) = \eta \cdot s(g) 
	\label{eq:hl_score}
\end{equation}

Where the score $s$ is defined as a function on the number of observed nodes and the missed ones:

\begin{equation}
	s(g) = observed \cdot  (1 - missed)
	\label{eq:score}
\end{equation}

And $\eta$ is a normalization factor calculated as:

\begin{equation}
	\eta = \frac{1}{\sum_{g \in G}^{} s(g)}
	\label{eq:eta}
\end{equation}

Equation \ref{eq:score} penalizes the explanations that contain missed nodes, which are nodes that should be present in the description but don't appear in $\hat{\sigma}$. This process is explained graphically in Figure \ref{fig:score}.

\begin{algorithm}[t]
	\DontPrintSemicolon
	\SetAlgoLined
	\KwIn{A plan library $L$, a set of observations $\hat{\sigma}$}
	\KwOut{A set of ranked explanations $P$}
	Initialize $P$ with the unmarked plans $T$ in $L$
	
	\ForEach{ $\hat{\sigma_i} \in \hat{\sigma}$}{
		Initialize $P^\prime$ to empty
		
		\While{$P$ is not empty}{
			Pop $p$ from $P$
			
			\ForEach{unobserved node $n$ in $p$ named $\hat{\sigma_i}$}{
				Generate a copy $p\prime$ of $p$
				
				Mark $n$ as observed in $p\prime$
				
				\If{there are any unobserved nodes on the left of $n$ in $p\prime$}{
					Mark them as missed
				}
			
				Insert $p\prime$ in $P^\prime$
			}			
		}
		Insert $P^\prime$ in $P$
	}
	Calculate the score for each plan;
	
	\Return{The generated explanations $P$, ordered by score}
	\caption{Explanations from partial observations}
	\label{algo:scoring}
\end{algorithm}

The procedure to derive the best explanation from a set of observations is described in Algorithm \ref{algo:scoring}. The process dynamically generates a set of explanations, each of which accounts for every possible interpretation of the observed symbols. This means that Algorithm \ref{algo:scoring} is able to deal with missed observations, where the robot might for any reason not record some of the actions performed by the partner. The explanation with the highest score at the end of the computation is chosen as the inferred intention.

\subsection{Real-time Verification}
\label{sub:verification}

The Verification module represents the robot's cognitive common sense and is responsible for the correctness of the predictions formulated step-by-step by both the Low-Level and the High-Level. In its essence, it serves the purpose of filtering out predictions that might arise at any level of abstraction which do not constitute valid statements on the state of the world. For example, an action prediction such as ``the human picks up the table and places it in the oven'' might be a possible statement generated by the Low-Level, but it makes no logical sense and must be a product of noise or a transient state.

In order to address the problem, we make use of a knowledge base in the form of an ontology in which we represent the entities of our world and the relations between them. Every time the Low-Level infers an action or the High-Level generates an explanation, these are verified through a Semantic Reasoner and if they are proven invalid they are discarded from the processing pipeline. Additionally, this component is also used by CASPER to decide which actions are assignable to the robot during the final collaborative decision-making. As an example, it would not be possible for the robot to ``eat the meal'', while on the contrary it would be capable of performing the action ``wash the dishes''.

On a technical level, CASPER makes use of an OWL2 ontology and a Pellet reasoner \cite{parsia2004pellet}.

\subsection{Collaborative Intelligence}
\label{sub:supervisor}

The final element of CASPER is the Supervisor, which is in charge of the coordination of the sub-processes that constitute the cognitive architecture. In particular, it manages the communication between these processes and collects the data produced by the Low-Level and High-Level in order to achieve the final purpose of this architecture: generate a collaborative behavior to help the partner with whatever tasks they are involved with.

To do so, it uses the goal explanation produced by the High-Level to generate an appropriate assistive plan. This involves the robot understanding which actions are yet to be executed and reason about which ones it is able to assist with. From the goal explanation, the Supervisor obtains the ``frontier'', that is the ordered set of all the unobserved nodes: these are the actions that have yet to be performed in order to achieve the goal. Using the validation procedure explained in Section \ref{sub:verification}, it then identifies the longest sequence of actions which the robot itself is able to perform. It then continues observing, waiting for the partner to execute the rest of the plan up to the point where the collaboration will start and in that moment it will send instructions to the Robot Control.

\section{Experiments}
\label{sec:experiments}

\begin{figure}[t]
	\centering
	\includegraphics[width=\linewidth]{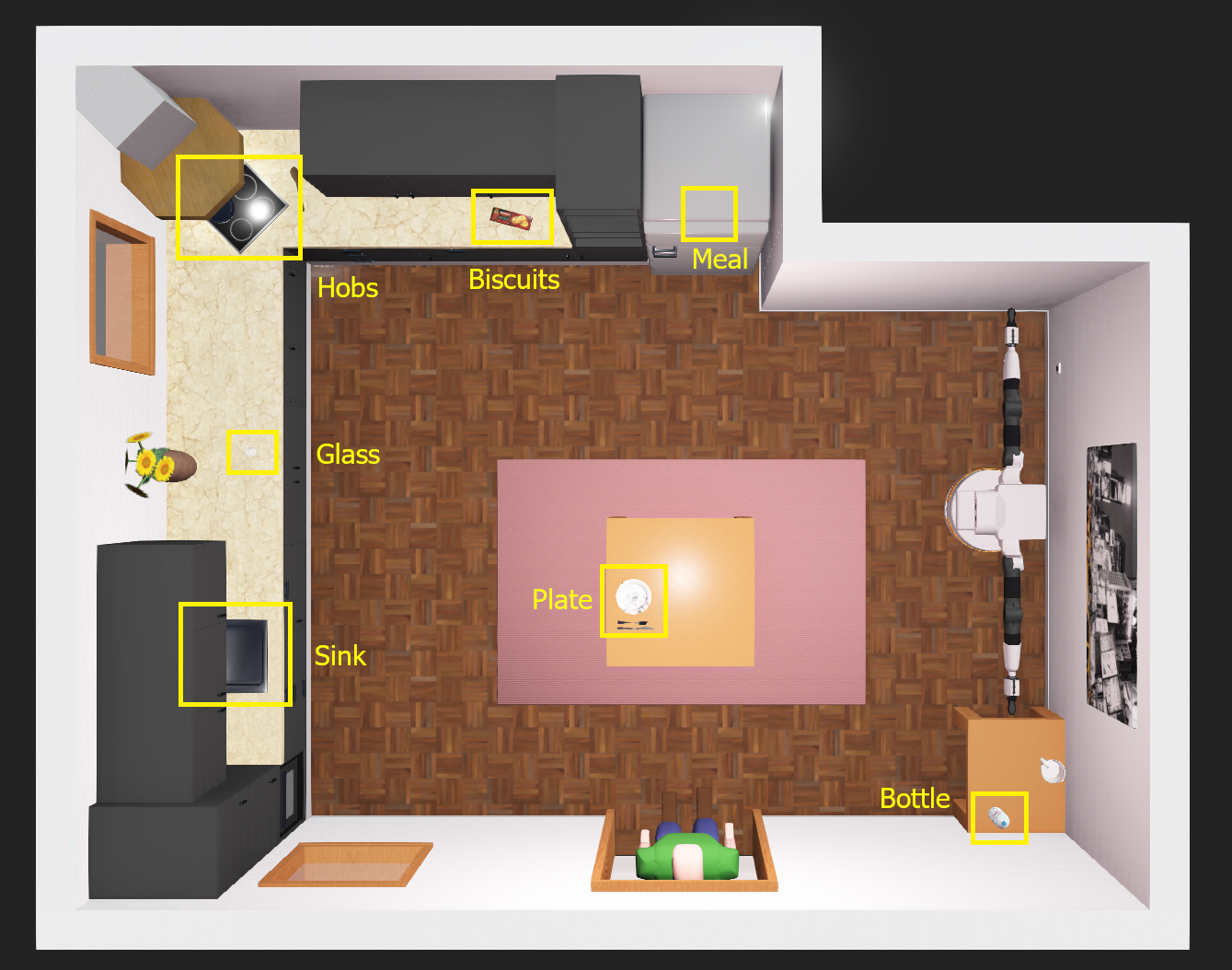}
	\caption{The experimental kitchen environment rendered in Webots. A TIAGo++ robot is tasked with observing a human performing actions in the scenery. The 7 OOIs are marked and annotated.}
	\label{fig:kitchen}
\end{figure}

\subsection{Experimental setup}
\label{sub:procedure}

In order to test the effectiveness of CASPER, we have decided to deploy it on a selected case study. In particular, we have developed an experiment involving a human and a TIAGo++ robot interacting inside a kitchen containing 7 OOIs: a water bottle, a canned meal, a box of biscuits, an empty glass, a plate, a set of hobs and a sink. This environment is shown in Figure \ref{fig:kitchen}. This was developed as a virtual environment created using the Webots open-source physics simulator \cite{michel2004cyberbotics}

The robot is instructed to visually find and track the human, collect observations on their actions within the room and process them through CASPER to infer their goal and generate an appropriate collaborative behavior. Both the human and the robot are able to navigate the environment, grasp items and release them.

\subsection{Low-Level}
\label{sub:exp_ll}

The \emph{movements} we have defined for our experiment are the following:

\begin{itemize}
	\item \textbf{STILL}: the agent is fixed in space, neither moving or interacting.
	\item \textbf{WALK}: the agent moves in the environment, holding nothing.
	\item \textbf{TRANSPORT}: same as WALK, but performed while carrying an object.
	\item \textbf{PICK}: the agent collects and item.
	\item \textbf{PLACE}: the agent positions a previously collected object somewhere in the environment.
\end{itemize}

In order to train the Decision Tree to map a set of QSRs to these movements, we need to generate a dataset. We do this by positioning the human in a random position inside the room and tasking them to stay still for a while, then walk to a random OOI, pick it up, transport it to a random destination, place it and then stand still again. This demonstrates the full range of movements that we wish to learn through our model. The robot, in turn, will observe the scene, calculate the QSRs and associate them to a label which is manually provided by the experimenter. A Decision Tree is a small-data model so we don't require a large amount of training samples: we repeat the previous procedure 10 times, then fit the model with the collected data.\\

\begin{figure}[htp]
	\subfloat[]{%
		\includegraphics[clip,width=\columnwidth]{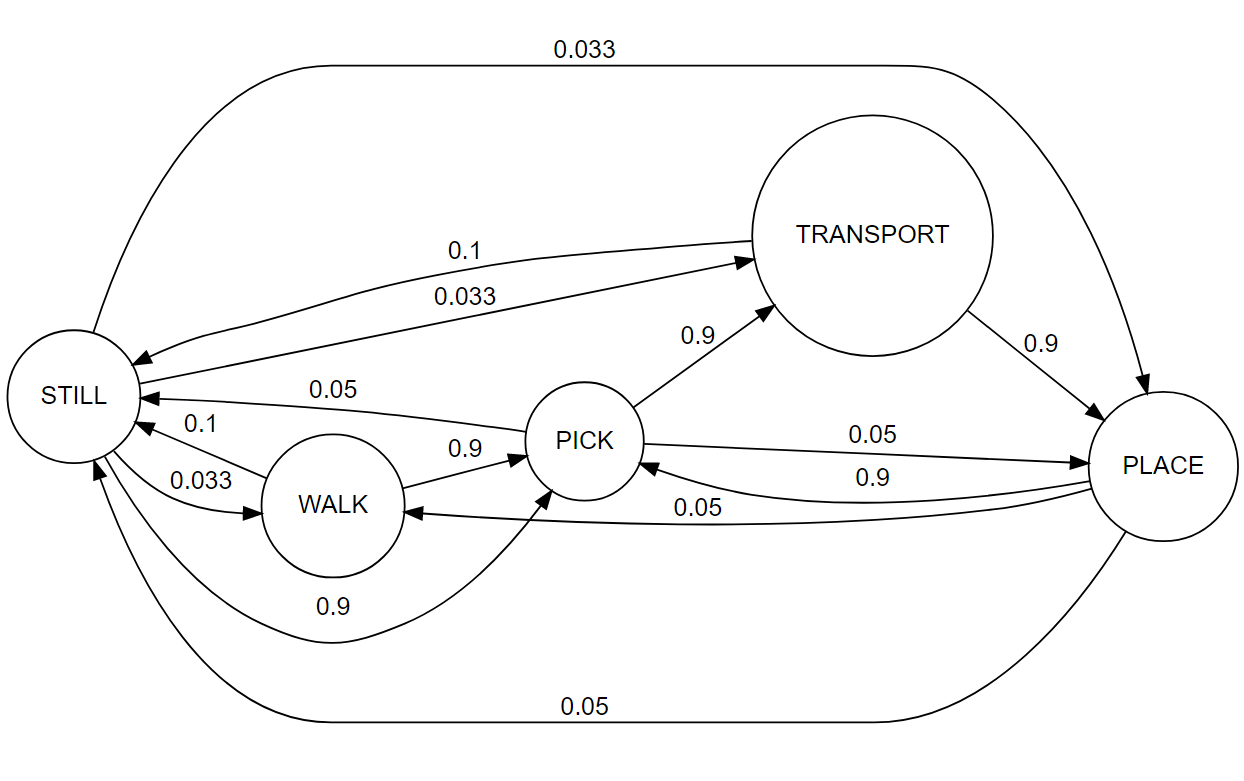}%
	}
	
	\subfloat[]{%
		\includegraphics[clip,width=\columnwidth]{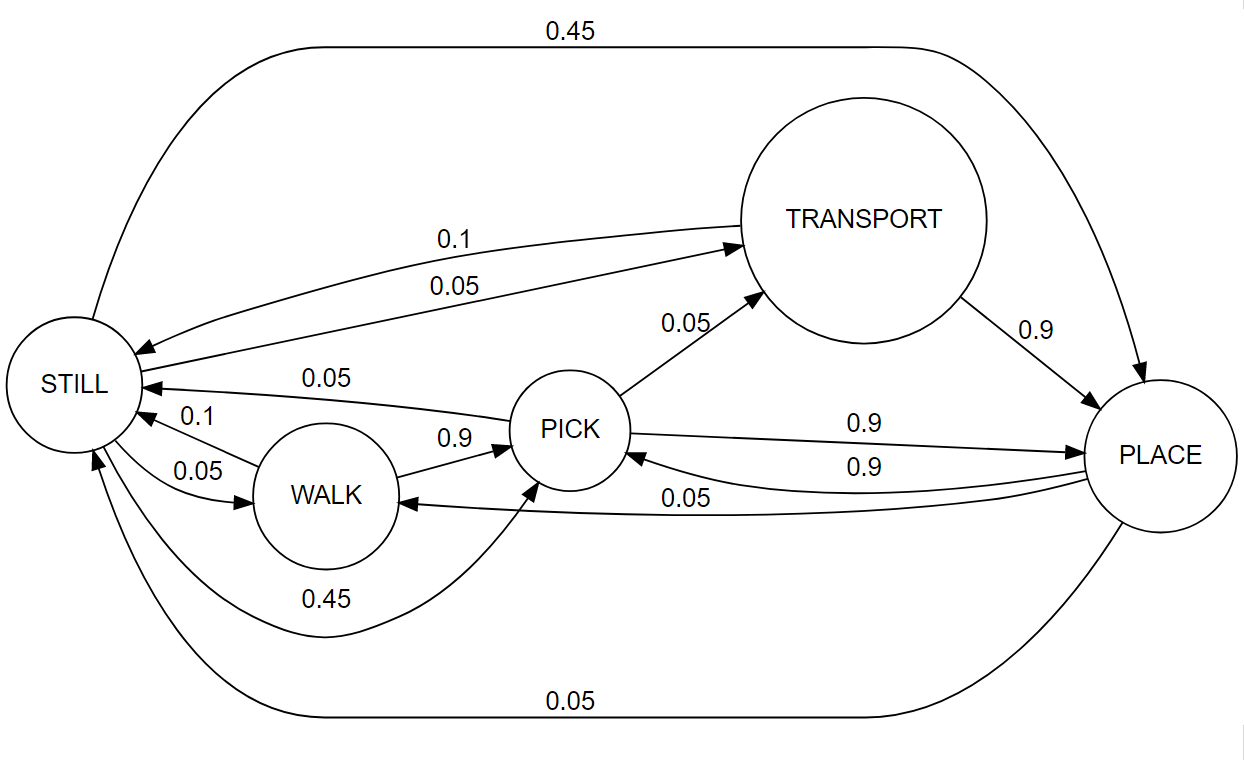}%
	}
	
	\subfloat[]{%
		\includegraphics[clip,width=\columnwidth]{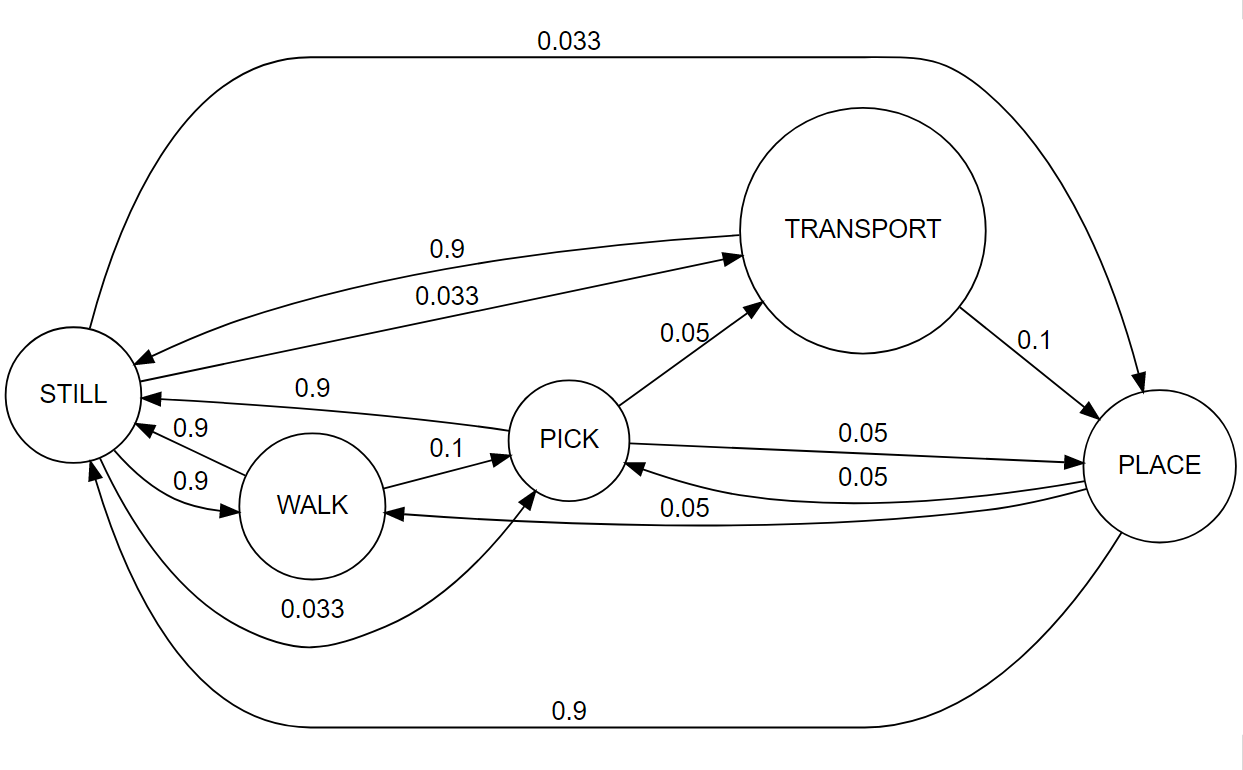}%
	}
	
	\caption{The FSMs which describe how each action is composed from the primitive movements: (a) Pick and Place, (b) Use, (c) Relocate.}
	\label{fig:fsm}
\end{figure}

The chosen \emph{actions} for the kitchen setup are:

\begin{itemize}
	\item \textbf{Pick and place}: a PICK movement followed by a TRANSPORT and terminated by a PLACE.
	\item \textbf{Use}: a loop of PICK and PLACE movements.
	\item \textbf{Relocate}: STILL, followed by WALK and another STILL movement.
\end{itemize}

The FSMs that describe these actions are reported in Figure \ref{fig:fsm}.

There is one action which requires contextualization: Use. This is because the latter can have a different meaning based on the location in which it is performed, or in other words the destination of the action. The lookup table that we use to contextualize it is reported in Table \ref{tab:context}.

\begin{table}[]
	\centering
	\caption{Contextualization for the action `Use'.}
	\label{tab:context}
	\begin{tabular}{|c|c|}
		\hline
		\textbf{Destination} & \textbf{Contextualized Action} \\ \hline
		Sink                 & Wash                           \\ \hline
		Hobs                 & Cook                           \\ \hline
		Plate                & Eat                            \\ \hline
		Glass                & Sip                            \\ \hline
	\end{tabular}
\end{table}

\subsection{High-Level}
\label{sub:exp_hl}

\begin{figure}[htp]
	\subfloat[]{%
		\includegraphics[clip,width=\columnwidth]{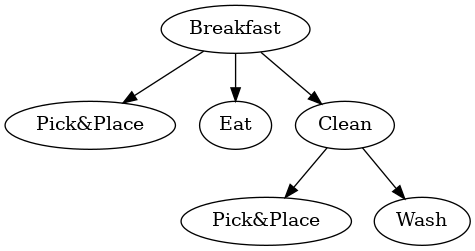}%
	}
	
	\subfloat[]{%
		\includegraphics[clip,width=\columnwidth]{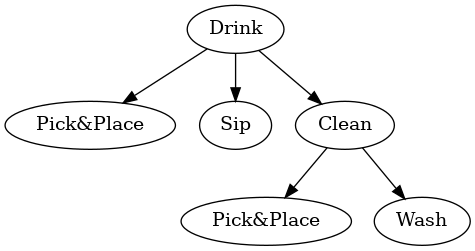}%
	}
	
	\subfloat[]{%
		\includegraphics[clip,width=\columnwidth]{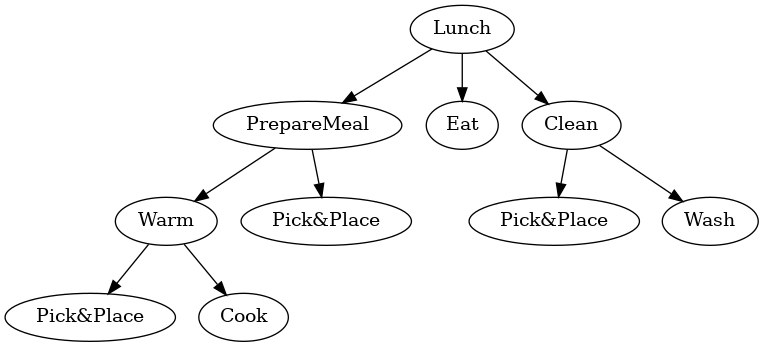}%
	}
	
	\caption{Detailed plans for each of the goals of the kitchen experiment: (a) Breakfast, (b) Drink, (c) Lunch.}
	\label{fig:goals}
\end{figure}

We define 3 distinct goals for our experiment:

\begin{itemize}
	\item \textbf{Breakfast}: the human will collect the biscuits, bring them to the plate and eat them, then move the plate to the sink and rinse it.
	\item \textbf{Drink}: the human will fetch the bottle of water, bring it to the glass and have a sip, then wash the glass.
	\item \textbf{Lunch}: the human will walk to the fridge and collect the canned meal, place it on the hobs and cook it. Afterwards, they will bring it to the plate, eat and wash the dishes.
\end{itemize}

The detailed plans for each of these goals are depicted in Figure \ref{fig:goals}.

In our plan library, as defined in Section \ref{sub:highlevel}:

\begin{align}
	\label{eq:sigma}
	\Sigma = [PickAndPlace, Wash, Cook, Eat, Sip] \\
	\label{eq:nt}
	NT = [PrepareMeal, Warm, Clean]
\end{align}

Note that the action Relocate is missing from Equation \ref{eq:sigma}: this is because our current experiment involves a single room. We have nevertheless implemented this action because of our future development plans for CASPER (see Section \ref{sec:conclusion}).

\subsection{Verification}
\label{sub:exp_v}

\begin{figure}[t]
	\centering
	\includegraphics[width=\linewidth]{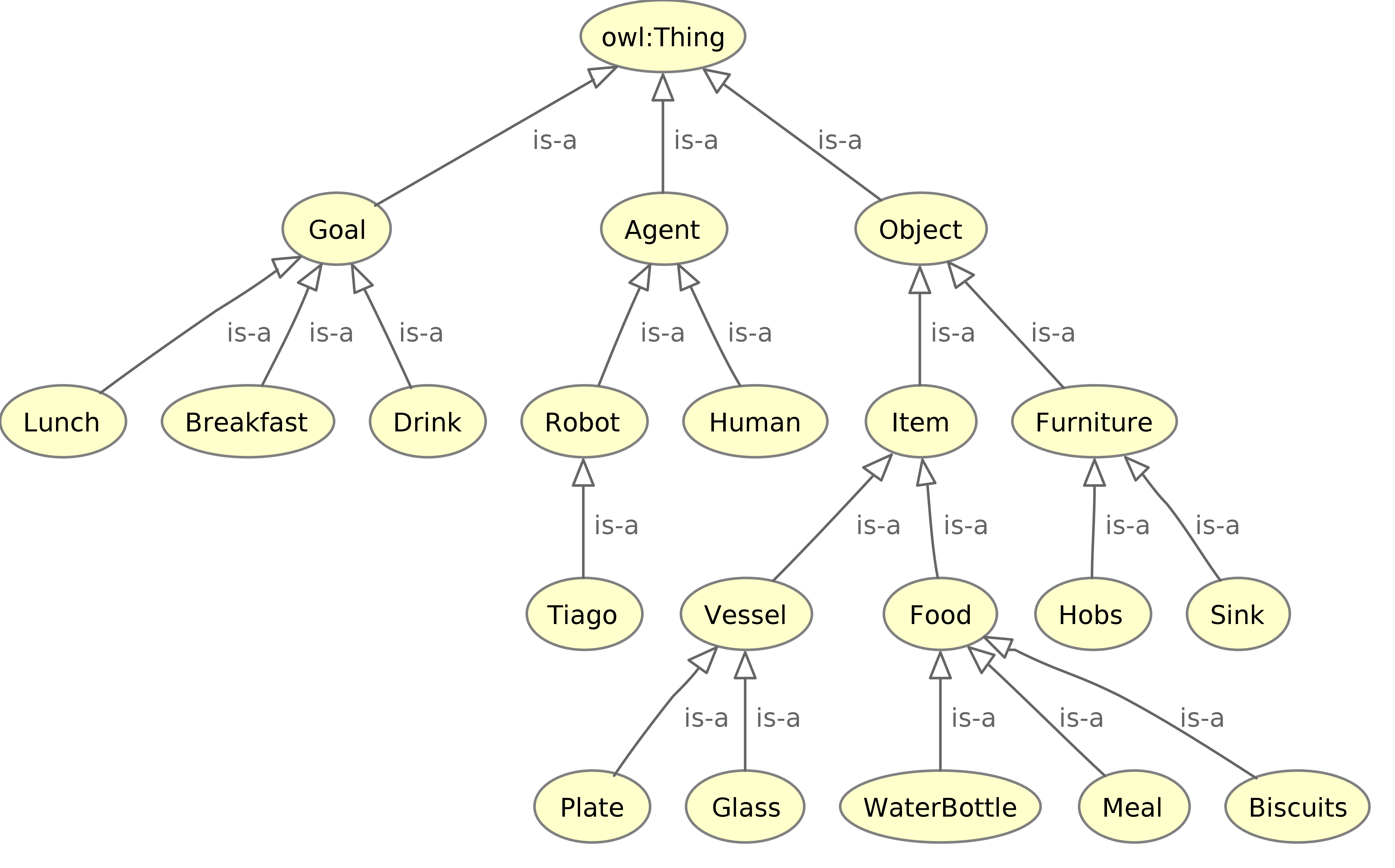}
	\caption{The ontology used as a knowledge base to perform verification during the kitchen experiment.}
	\label{fig:onto}
\end{figure}

The ontology which we use to describe the kitchen experiment is reported in Figure \ref{fig:onto}. This knowledge base defines each entity in the environment as belonging to one of three macro-groups: Goals, Agents or Objects. Each Agent can be a Human or a Robot, the latter only containing the TIAGo++ robot we are using but potentially expandable to include several kinds of robots grouped by their capabilities (for example, humanoid and non-humanoid). The Objects can be Items or Furniture: the former includes both vessels and food.

\begin{table}[]
	\centering
	\caption{Data properties for the ontological representation of the kitchen experimental environment.}
	\label{tab:props}
	\begin{tabular}{c|c|c|c|c|c|}
		\cline{2-6}
		& Move & Eat & Drink & Cook & Wash \\ \hline
		\multicolumn{1}{|c|}{Sink}        &      &     &       &      &      \\ \hline
		\multicolumn{1}{|c|}{Hobs}        &      &     &       &      &      \\ \hline
		\multicolumn{1}{|c|}{Plate}       & \checkmark  &     &       &      & \checkmark  \\ \hline
		\multicolumn{1}{|c|}{Glass}       & \checkmark  &     & \checkmark   &      & \checkmark  \\ \hline
		\multicolumn{1}{|c|}{Biscuits}    & \checkmark  & \checkmark &       &      &      \\ \hline
		\multicolumn{1}{|c|}{Meal}        & \checkmark  & \checkmark &       & \checkmark  &      \\ \hline
		\multicolumn{1}{|c|}{WaterBottle} & \checkmark  &     & \checkmark   &      &      \\ \hline
	\end{tabular}
\end{table}

Each element is characterized by some properties which define the kind of interactions that are possible with each of them. These are reported in Table \ref{tab:props}. For example, the biscuits can be eaten and moved, but not cooked, drinked or washed. At the same time, the ontology defines some object properties, i.e. relations between entities of the knowledge base which impose limitations useful to verify the validity of the statements produced by the Low-Level and the High-Level. For example, the Eat action might only be performed by Humans with a target which is eatable and a destination which is a Vessel. A pair of object properties are defined for each of the actions (including the contextualized ones).

Finally, we use Semantic Web Rule Language (SWRL) \cite{horrocks2004swrl} definitions to allow the reasoner to perform inferences on incomplete statements. For example, we know that if an Agent is washing an item, then the destination must be a Sink.

\section{Results and Discussion}
\label{sec:results}

\subsection{Focus Evolution in Time}

\begin{figure*}[t]
	\centering
	\includegraphics[width=.85\textwidth]{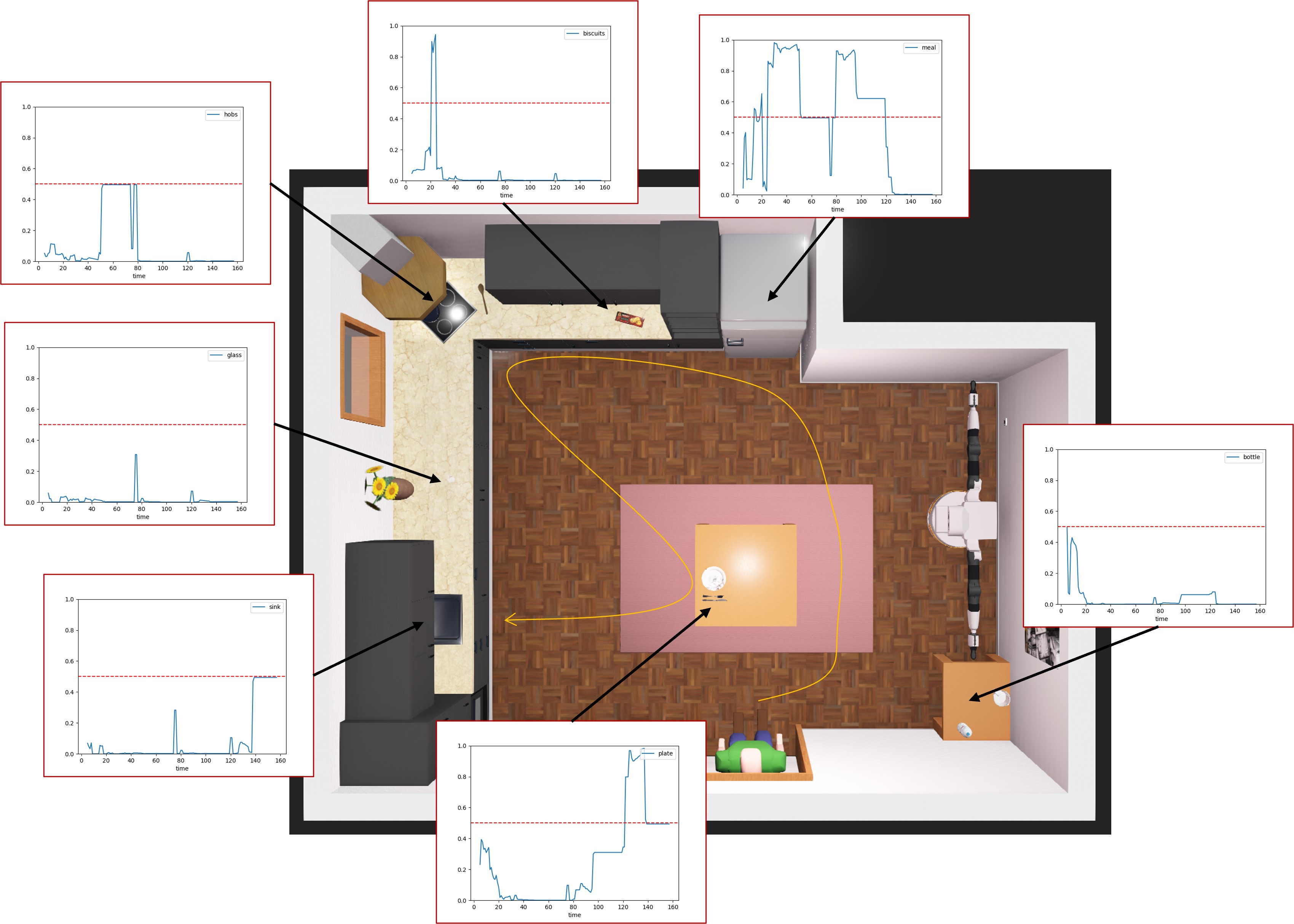}
	\caption{Each OOI is annotated with a graph showing the evolution of the focus estimation probability in time during the execution of the goal `Lunch'. The arrow on the floor shows the trajectory of the human through the environment.}
	\label{fig:focus_evolution}
\end{figure*}

Figure \ref{fig:focus_evolution} explores the output of the Focus Estimator while observing a human performing the goal Lunch (which, we recall, is executed by collecting the meal from the fridge, cooking it on the hobs, eating it at the table and finally washing the dishes). In particular, each OOI $o_i$ is annotated with a graph describing the temporal evolution of the assigned normalized probability values $P(o_i)$. Of course, at each timestep: $\sum_{o_i \in O} P(o_i) = 1$.

At the start of the simulation, the farthest OOIs have the lowest chance to be considered as the target of the observed agent's attention. When the human starts moving and turns to their right, the probability for the Bottle increases up to the designed threshold of 0.5, but soon after they face away and the score drops. Thanks to the sliding window, this OOI is not realistically considered as the human's target. 

The agent then continues its path to the north: the Plate's probability decreases as it exits their field of view, while the probabilities for the Meal increases steadily. Around timestep 20, the human turns momentarily towards the Biscuits and once again creates a probability spike which does not last long enough to influence the system. Once the Meal is grasped, $QDC(Meal) = Touch$ and $QTC(Meal) = 0$, so the focus estimation for this OOI is high.

Further ahead in time, around timestep 50, the focus is evenly divided between the Meal and the Hobs during the Cook action. The human then turns towards the table: the probability for the Hobs decreases rapidly while the ones for the Plate increase. At this point, around timestep 90, the human finishes approaching the table and starts eating. The Focus Estimation divides more or less equally the probability for both the Meal and the Plate, with a 60/40 split.

Finally, the Plate is brought to the Sink. The focus score for the former is high and eventually evens out with the latter, while the probability value of the Meal drops to 0 very quickly.

Overall, Figure \ref{fig:focus_evolution} shows that the Focus Estimator module is correctly able to predict the human's attention while they move around the environment.

\subsection{Decision Tree Training}

\begin{figure*}[t]
	\centering
	\includegraphics[width=.6\textwidth]{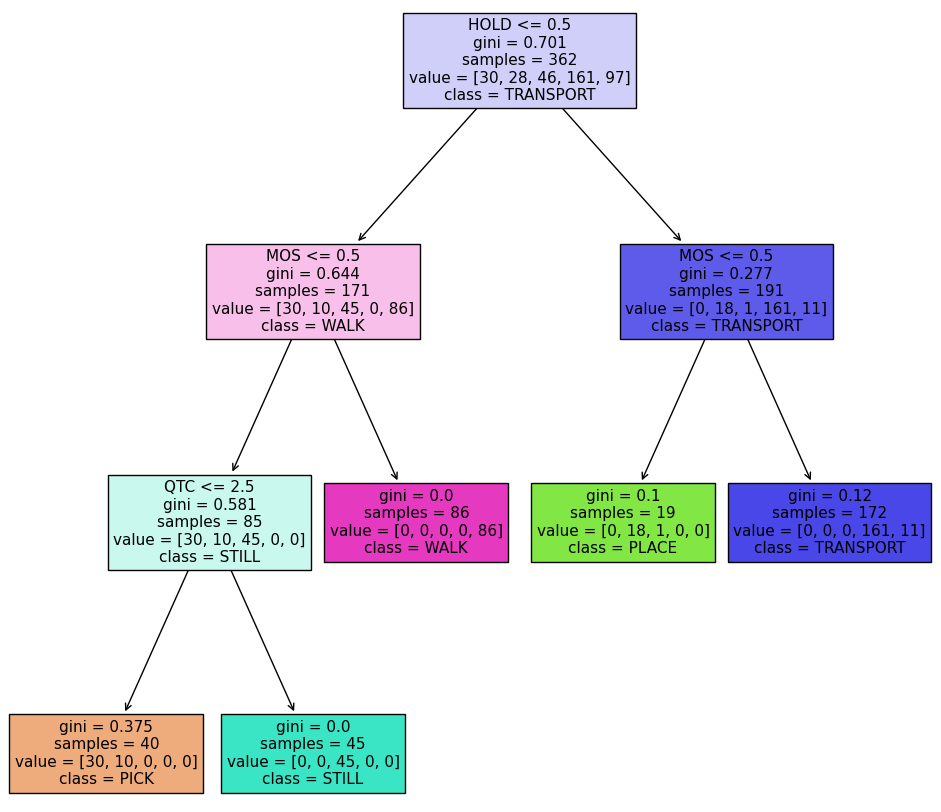}
	\caption{The Decision Tree trained from the experimental data in the kitchen collaboration environment. The QSRs are classified as one of the following movements: Still, Walk, Transport, Pick or Place.}
	\label{fig:decision_tree}
\end{figure*}

The Decision Tree that acts as the Movement Predictor is trained on a dataset generated in the simulated environment following the procedure described in Section \ref{sub:exp_ll}. This dataset contains 362 training examples obtained from 10 random trials. The total number of samples per class is the following: STILL (46), WALK (97), PICK (30), TRANSPORT (161) and PLACE (28). The model fitted on this data is shown in Figure \ref{fig:decision_tree}.

To evaluate the fitness of this model, we have performed a 10-fold cross-validation: we split our dataset into the 10 groups from which it was generated, leave one aside for testing and train on the remaining, then repeat for each of the unique groups. Our average 10-fold cross-validation accuracy is 0.94.

\subsection{Markov Chain Finite-State Machines}

\begin{figure}[htp]
	\centering
	\subfloat[]{%
		\includegraphics[clip,width=.9\columnwidth]{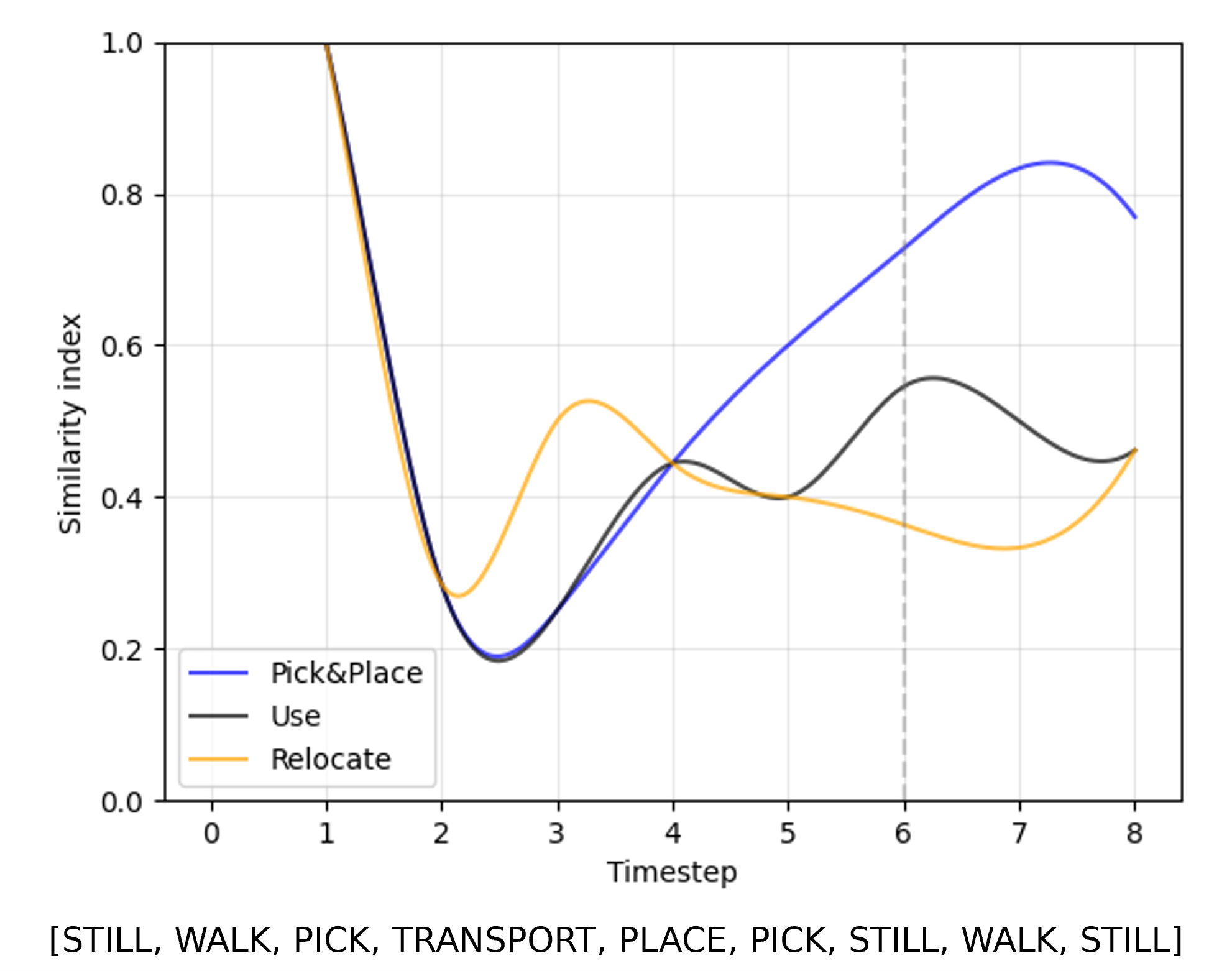}%
	}
	
	\subfloat[]{%
		\includegraphics[clip,width=.9\columnwidth]{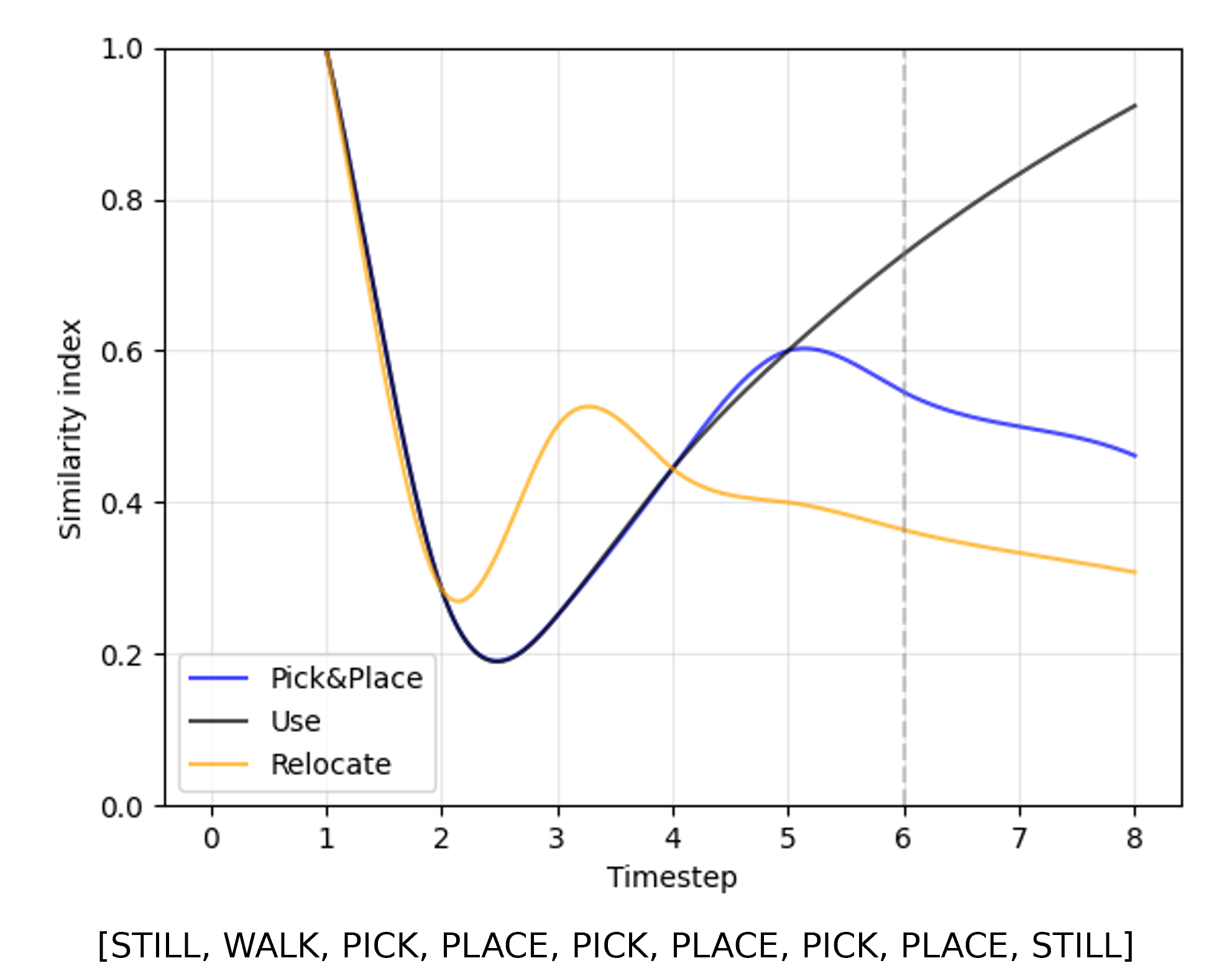}%
	}
	
	\subfloat[]{%
		\includegraphics[clip,width=.9\columnwidth]{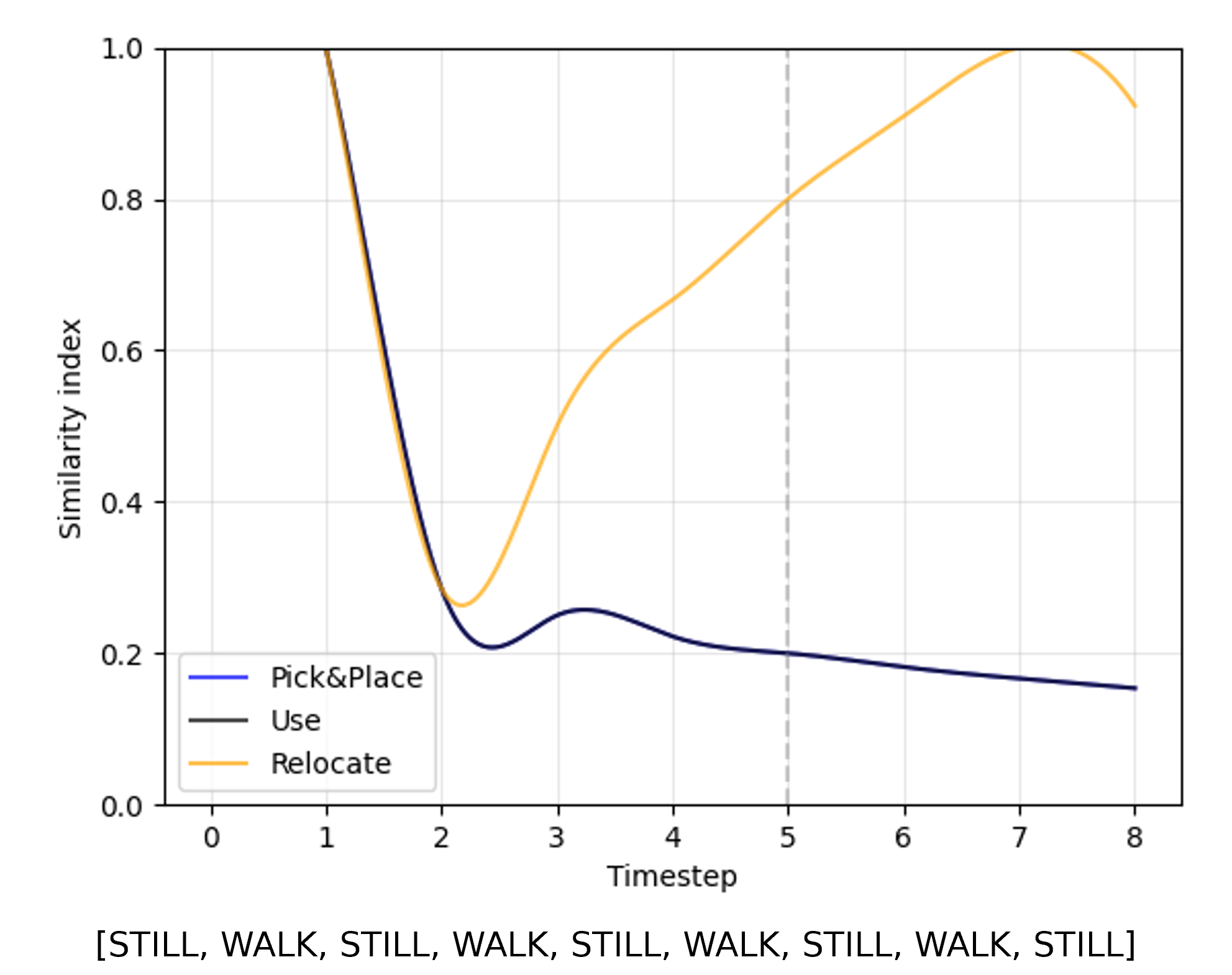}%
	}
	
	\caption{Temporal response of the three FSMs which define the actions `Pick and Place', `Use' and `Relocate' on different sequences of movements (reported below each graph). The vertical dashed line indicates the moment in which the ensemble has inferred the action.}
	\label{fig:ensemble}
\end{figure}

To evaluate the fitness of the Action Predictor, we generate 3 sequences of 9 movements and we input them incrementally into the ensemble to analyze its temporal response to the observations. The results are reported in Figure \ref{fig:ensemble}, where we have plotted the similarity score calculated through the Ratcliff-Obershelp algorithm at each timestep.

Since the sampling of each FSM uses the initial observed symbol as the starting state, the similarity score is maximum for the first iterations. Despite that, none of the models prevails on the others and no winner is declared yet. As soon as more symbols are fed into the system, the scores start to oscillate and differ, leaving one clear winner: this is the model that best describes the observed sequence of observations. The first two sequences produce a prediction on timestep 6, while the third one receives an inference on timestep 5. These predictions are in line with what we would expect given the symbols in the input sequences.

\subsection{Explanation Generation}
\label{sub:results_hl}

\begin{table*}[]
	\centering
	\caption{Experimental results on the Goal Reasoner. See Section \ref{sub:results_hl} for the full explanation.}
	\label{tab:pl_results}
	\begin{tabular}{|c|c|c|c|c|c|}
		\hline
		\textbf{TRIAL}       & \textbf{ACTIONS} & \textbf{EXPLANATIONS} & \textbf{TIME ($\mu s$)} & \textbf{CONFIDENCE} & \textbf{OUTCOME} \\ \hline
		\multirow{2}{*}{\#1} & Pick\&Place      & 7                     & 3.58          & 0.23                &                  \\ \cline{2-6} 
		& Eat              & 3                     & 1.91          & 0.53                & Breakfast        \\ \hline
		\multirow{2}{*}{\#2} & Pick\&Place      & 7                     & 5.72          & 0.23                &                  \\ \cline{2-6} 
		& Sip              & 1                     & 2.15          & 1                   & Drink            \\ \hline
		\multirow{2}{*}{\#3} & Pick\&Place      & 7                     & 3.34          & 0.23                &                  \\ \cline{2-6} 
		& Cook             & 1                     & 1.91          & 1                   & Lunch            \\ \hline
		\multirow{3}{*}{\#4} & Pick\&Place      & 7                     & 3.58          & 0.23                &                  \\ \cline{2-6} 
		& Pick\&Place      & 5                     & 3.34          & 0.28                &                  \\ \cline{2-6} 
		& Eat              & 1                     & 1.43          & 1                   & Lunch            \\ \hline
		\multirow{3}{*}{\#5} & Pick\&Place      & 7                     & 4.29          & 0.23                &                  \\ \cline{2-6} 
		& Pick\&Place      & 5                     & 3.34          & 0.28                &                  \\ \cline{2-6} 
		& Pick\&Place      & 1                     & 1.19          & 1                   & Lunch            \\ \hline
		\multirow{3}{*}{\#6} & Pick\&Place      & 7                     & 3.34          & 0.23                &                  \\ \cline{2-6} 
		& Pick\&Place      & 5                     & 2.62          & 0.28                &                  \\ \cline{2-6} 
		& Wash             & 5                     & 2.62          & 0.3                 &                  \\ \hline
		\#7                  & Sip              & 1                     & 3.1           & 1                   & Drink            \\ \hline
		\#8                  & Eat              & 2                     & 2.38          & 0.69                &                  \\ \hline
		\multirow{2}{*}{\#9} & Wash             & 3                     & 4.29          & 0.41                &                  \\ \cline{2-6} 
		& Cook             & 0                     & 4.53          & 0                   &                  \\ \hline
	\end{tabular}
\end{table*}

To test the Goal Reasoner embedded in the High-Level module, we have run 9 trials in which we have provided it with several sequences of observations. For the rest of this discussion, please refer to Figure \ref{fig:goals} for the structure of our Goal Library.

Table \ref{tab:pl_results} summarizes the data we have collected. The latter shows, from left to right: the id of the trial, the observation (action) that was incrementally input in the system, the number of explanations generated by the reasoner, the time in microseconds required to produce the result, the confidence of the top-scoring explanation and finally the output of the component. If the latter is blank, then the reasoner could not formulate a prediction, otherwise it will report the name of the goal whose plan explains the observations. The system was reset between each trial.

Trial 1 presents to the High-Level the actions `Pick and place' and `Eat'. There are three possible explanations for these observations: one that describes the goal Breakfast with only observed and unobserved nodes and two that represent the plan for Lunch that accounts for several missing nodes (recall that an unobserved node is marked as missed if another node is observed on its right-hand side). Since our model assigns a higher score to the simplest model to describe the data, Breakfast is chosen as the prediction.

Trials 2 and 3 are straightforward: the goals Drink and Lunch are the only ones that contain respectively the actions `Pick and Place' followed by either `Sip' and `Cook', so the Goal Reasoner can formulate a very confident prediction.

Trial 4 and 5 are more ambiguous: both of them begin with two `Pick and place' actions which could describe each of the models with similar probability. Only when the system receives the third observation it can commit to a clear inference. Trial 6 follows a similar narrative, but the third input is still not able to disambuigate the goal: each of the possible plans share an uniform probability distribution and the reasoner fails to produce a prediction. No additional inputs would change the situation, since the rightmost node of the tree has been observed (remember that we assume that each observation happens after the preceding one). Of course, one could object that the goal could indeed by identified by the OOIs with which the human is interacting. This will be done in the full-scale experiment by the Verification component, which uses its ontology to filter out invalid explanations such as the goal Lunch if the first action `Pick and Place' has been performed on the Biscuits.

In trial 7, we provide the system with a single observation that, alone, is able to discern the goal. We try doing the same with the observation `Eat' in trial 8, but that action on its own could describe both Breakfast or Lunch. No further observation would be able to disambiguate this scenario, since the only next possible actions are `Pick and Place' and `Wash', which are common to both the candidates.

Finally, trial 9 shows an invalid sequence of actions: there is no goal plan in which the action `Wash' is followed by any other action. For this reason, the system produced no valid explanations.

\subsection{Intention Reading and Collaboration}
\label{sub:ci}

\begin{table}[]
	\centering
	\caption{CASPER's performance on the kitchen experiment. See Section \ref{sub:ci} for the full explanation.}
	\label{tab:results}
	\resizebox{\columnwidth}{!}{%
		\begin{tabular}{|c|c|c|c|c|c|c|}
			\hline
			\textbf{Goal} & \textbf{Observed} & \textbf{Missed} & \textbf{Waiting} & \textbf{Planned} & \textbf{Accuracy} & \textbf{Time (s)} \\ \hline
			Breakfast & 1.0 & 0.2 & 0.0 & 1.8 & 100\% & 42.21 \\ \hline
			Drink     & 1.0 & 0.4 & 0.0 & 1.6 & 100\% & 54.14 \\ \hline
			Lunch     & 1.8 & 0.2 & 1.8 & 2.0 & 100\% & 82.46 \\ \hline
		\end{tabular}%
	}
\end{table}

Having verified the single components that constitute the Low-Level and High-Level modules of CASPER, we are now ready to analyze the overall performance of the cognitive architecture working together to read the human's intention and producing collaborative decision-making. The procedure we have followed is the following: we have run 5 trials for each of the 3 goals, randomizing the human's starting position and collecting a total of 15 data samples. For each of these, we have recorded: the number of observed and missed nodes in the winning explanation, the number of actions that the robot waits for the partner to complete before collaborating, the number of actions that the robots plans to execute, the accuracy of the prediction and the time, in seconds, needed to make an inference. The mean values of these variables are collected in Table \ref{tab:results}.

The first thing to notice is that the robot was always able to correctly read the human's intention, despite some noise in the perceptual data collected by the system: the synergistic interaction of CASPER's components results in a robust intention reading performance. Table \ref{tab:results} also indicates that the time and observations required to infer the goal Lunch were higher than for the other two goals: this is in line with the higher complexity of its plan compared to the ones for Breakfast and Lunch.

The collaborative plan calculated from the cognitive architecture in each case was to wait for the human to finish eating their meal and then clean up the kitchen, which involves transporting the plate to the sink and washing it.

\subsection{Verification}

\begin{figure}[t]
	\centering
	\includegraphics[width=\linewidth]{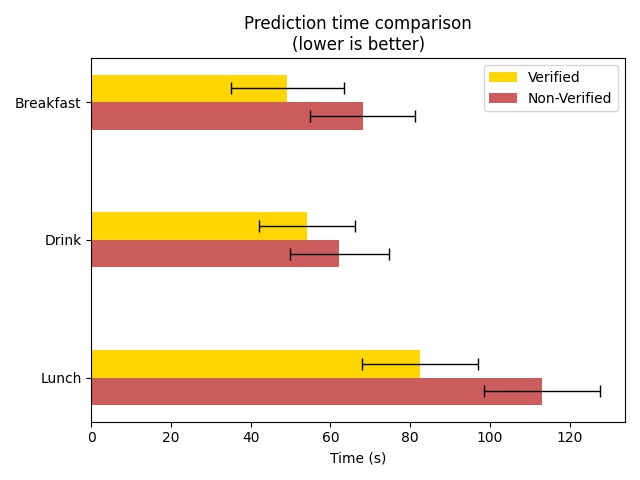}
	\caption{Average prediction time of CASPER with and without the Verification module. Despite the computational overhead of the Pellet reasoner, the verification module ensures a faster inference from less observations.}
	\label{fig:comparison}
\end{figure}

An additional experiment was carried out to investigate the performance of the Verification module. In particular, we run the same experiment used to generate the data we have discussed in Section \ref{sub:ci}, this time disabling the formal verification of both the Low-Level and the High-Level: we call this the Non-Verified condition, as opposed as the Verified condition which represents the full cognitive architecture.

Figure \ref{fig:comparison} shows the comparison between the two configurations. The results that we collected prove that by disabling the Verification module we don't hurt the accuracy of the system, but instead we cause a slower prediction time. Despite the computational overhead introduced by the semantic reasoner, the Verified condition outperformed the Non-Verified one. This happens because the latter requires more observations to make sense of the environment, whilst the Verification module is able to discard noisy observations and illogical explanations before either of them are further processed, cutting down the overall inference time.

\section{Conclusion and Future Work}
\label{sec:conclusion}

In this paper, we have introduced CASPER: a symbolic cognitive architecture designed to perform intention reading and to calculate collaborative behaviors for human-robot teaming scenarios. Our system is able to accomplish the task through a set of parallel processes that communicate with each other and that can translate QSR descriptors into movements, then into actions and finally into goals and sub-goals using a bottom-up approach. Through the implementation of a simulated experimental case study based on a kitchen environment, we have proven the soundness of our methodologies.

The design of this system is driven by the requirement to embed in social robots the ability to autonomously integrate themselves in the structure of our daily routines, without the need for a human operator to explicitly provide instructions for the machine. Instead, by being able to understand the actions of other agents within the environment, a robot endowed with this cognitive architecture is able to seamlessly cooperate with them. In fact, despite our focus on human-robot interaction, this architecture would be equally applicable to robot-robot interactions.

Our main scientific contribution is the demonstration that QSRs can be used as an efficient means to achieve intention reading capabilities in artificial intelligence systems, a proof of concept that is lacking in the current state-of-the-art. Our technological contribution comes in the form of a cognitive architecture that incorporates novel algorithms for perception, reasoning and action selection which take inspiration from psychology and cognitive science.

According to the taxonomy defined by Kotseruba and Tsotsos \cite{kotseruba202040}, CASPER is a symbolic architecture which implements the most common cognitive mechanisms, which we shall now summarize. Perception, the process that transforms raw input into the system’s internal representation for carrying out cognitive tasks, is performed by vision. Attention is modeled as a viewpoint/gaze selection mechanism: this means that the robot endowed with CASPER is able to select a target and track it through space. Action selection, which drives decision-making, is the result of a planning strategy which aims to maximize the relevance of the selected behavior. This architecture possesses all three types of memory: sensory (the QSR Library), working (handled by the Low- and High-Level) and long-term (in the form of the learned models and ontology). Learning comes in the form of declarative knowledge, which is a collection of facts about the world and various relationships defined between them. Reasoning is a cognitive ability which is present and central to each and every cognitive architecture, including this one. Finally, CASPER also implements metacognition, that is the ability to reason about one's own thoughts: this is done by the Verification module, which constantly monitors the other internal processes, identifying and correcting any erroneous decisions.

A limitation of our work is represented by the intrinsic nature of symbolic artificial intelligence: our methodologies suffer from the Knowledge Acquisition Bottleneck \cite{cullen1988knowledge}, which refers to the human intervention required to translate real-world conditions in symbolic inputs for the intelligent system. In our case, this comes in the form of the selection of OOIs, the plans for each goal and the ontology that envelops the properties of the environment. Despite this, we argue that this disadvantage is compensated by the interpretability of each of the components that build up CASPER (the decisions of which can be explained at each step of computation) and the lack of computationally expensive and data-hungry processes.

Another limitation is given by the fact that we have tested CASPER in simulation and with fairly simple goals. The reason for this is that the work presented here is foundational to the true purpose of this cognitive architecture: that is, to offer support for intention reading and trust considerations in heterogeneous multi-agent teaming scenarios. Our planned future work involves the creation of a more complex environment composed of multiple rooms where distributed groups of humans will be interacting with robots of different make and capabilities in order to achieve a Team Goal. In this kind of setting, the individual robots will not be able to gather all the necessary information needed to predict the shared objective, rather they will have to rely on partial observations and communication with their peers. Moreover, performing empirical trials in simulation gives us the freedom to experiment with arbitrarily different environments, including varying numbers and type of agents.

One additional future expansion on CASPER is the inclusion of artificial trust considerations \cite{vinanzi2021collaborative, vinanzi2019trust, centeio2022artificial} through which the endowed robot will be able to assess the capabilities of other agents (humans or robots) to pursue the desired goal. Our hypothesis is that this cognitive skill will be valuable in allowing the group of heterogeneous robots to sort each other between the several available collaborative assignments required to assist the humans of their team.

\bibliographystyle{unsrt}
{\footnotesize \bibliography{biblio.bib}}

\end{document}